\documentclass{article} 
\usepackage{iclr2027_conference,times}

\usepackage[utf8]{inputenc} 
\usepackage[T1]{fontenc}    
\usepackage{hyperref}       
\usepackage{url}            
\usepackage{booktabs}       
\usepackage{amsmath}        
\usepackage{amssymb}        
\usepackage{amsfonts}       
\usepackage{microtype}      
\usepackage{xcolor}         
\usepackage{graphicx}
\usepackage{multirow}
\usepackage{colortbl}       
\usepackage{array}          
\usepackage[most]{tcolorbox} 
\usepackage{pgfplots}        
\pgfplotsset{compat=1.17}

\definecolor{headerbg}{RGB}{235,240,247}      
\definecolor{ourshade}{RGB}{255,243,224}      
\definecolor{ablrow}{RGB}{248,248,248}        
\definecolor{ctrlrow}{RGB}{238,244,236}       
\definecolor{rulecolor}{RGB}{40,40,40}        
\definecolor{groupclr}{RGB}{55,80,120}        
\definecolor{accentclr}{RGB}{180,80,40}       

\colorlet{rowblue}{ctrlrow}
\colorlet{rowgray}{headerbg}
\colorlet{ourpink}{ourshade}

\definecolor{promptframe}{RGB}{60,90,150}
\definecolor{promptback}{RGB}{240,245,255}
\newcommand{\up}[1]{\textcolor{green!55!black}{\scriptsize$\uparrow$#1}}

\newcommand{\ours}[1]{\textbf{\textcolor{accentclr}{#1}}}
\newcommand{\groupheader}[2]{\rowcolor{headerbg}\multicolumn{#1}{l}{\textit{\textcolor{groupclr}{#2}}}}
\newcommand{\thickrule}{\noalign{\hrule height 0.9pt}}
\newcommand{\medrule}{\noalign{\hrule height 0.5pt}}
\newtcolorbox{promptbox}[1][]{
  enhanced, breakable,
  colback=promptback, colframe=promptframe,
  boxrule=0.6pt, arc=2pt, left=6pt, right=6pt, top=4pt, bottom=4pt,
  fonttitle=\bfseries\small,
  coltitle=white,
  colbacktitle=promptframe,
  title={#1}
}
\definecolor{codeframe}{RGB}{80,80,80}
\definecolor{codeback}{RGB}{248,248,248}
\newtcolorbox{codebox}[1][]{
  enhanced, breakable,
  colback=codeback, colframe=codeframe,
  boxrule=0.5pt, arc=2pt, left=6pt, right=6pt, top=3pt, bottom=3pt,
  fonttitle=\bfseries\small,
  coltitle=white,
  colbacktitle=codeframe,
  title={#1}
}
\definecolor{exframe}{RGB}{120,80,140}
\definecolor{exback}{RGB}{248,244,252}
\newtcolorbox{examplebox}[1][]{
  enhanced, breakable,
  colback=exback, colframe=exframe,
  boxrule=0.5pt, arc=2pt, left=6pt, right=6pt, top=4pt, bottom=4pt,
  fonttitle=\bfseries\small,
  coltitle=white,
  colbacktitle=exframe,
  title={#1}
}

\title{Learning Spatiotemporal Sensitivity in Video LLMs via Counterfactual Reinforcement \\ Learning}

\makeatletter
\newcommand{\symfootnote}[1]{\begingroup\renewcommand\thefootnote{}\renewcommand\@makefnmark{}\footnotetext{#1}\endgroup}
\makeatother
\author{%
\bfseries Dazhao Du\textsuperscript{1,2,$\ast$},\enspace Jian Liu\textsuperscript{1},\enspace Jialong Qin\textsuperscript{1},\enspace Tao Han\textsuperscript{1},\enspace Bohai Gu\textsuperscript{1},\enspace Fangqi Zhu\textsuperscript{1},\\[2pt]
\bfseries Yujia Zhang\textsuperscript{2},\enspace Eric Liu\textsuperscript{2},\enspace Xi Chen\textsuperscript{2},\enspace Song Guo\textsuperscript{1,$\dagger$}\\[5pt]
{\normalfont\normalsize\textsuperscript{1}Hong Kong University of Science and Technology\qquad\textsuperscript{2}Tencent}%
}

\iclrfinalcopy
\begin{document}
\maketitle
\lhead{Preprint}
\symfootnote{$^{\ast}$Work done during an internship at Tencent.}
\symfootnote{$^{\dagger}$Corresponding author.}

\maketitle

\begin{abstract}
Video large language models (Video LLMs) can achieve strong video-QA accuracy without reliably tracking spatiotemporal dynamics. A model may answer a motion question from static cues, for example, and give the same prediction even after the underlying motion is reversed. Correctness-based reinforcement learning does not directly address this problem because it rewards the final answer without requiring the policy to respond to task-relevant video dynamics. We propose \textbf{Counterfactual Relational Policy Optimization (CRPO)}, which explicitly trains Video LLMs to respond to controlled changes in the visual input. For each training example, CRPO constructs a counterfactual video, such as a horizontally flipped or temporally reversed version of the original, and jointly optimizes rollouts from both videos under a shared policy. Factual supervision anchors \emph{what the model should answer}, while counterfactual supervision constrains \emph{when that answer should change}: predictions should change when an intervention alters task-relevant dynamics and remain stable when the queried property is preserved.
This coupling provides a direct behavioral learning signal without requiring ground-truth labels for transformed videos or annotated spatiotemporal reasoning traces, while discouraging indiscriminate answer changes.
To evaluate this property, we introduce \textbf{DyBench}, a paired counterfactual benchmark with 3{,}014 videos and a strict pair-accuracy metric. Across paired spatiotemporal evaluations and standard video benchmarks, CRPO improves sensitivity to motion and temporal changes while improving performance on general video understanding. The gains also extend to segment reordering, a transformation never used during training, suggesting that CRPO learns sensitivity to video dynamics beyond the training interventions. The project website can be found at \url{https://ddz16.github.io/crpo.github.io/}.
\end{abstract}

\section{Introduction}
\label{sec:intro}

\begin{figure}[t]
  \centering
  \includegraphics[width=\textwidth]{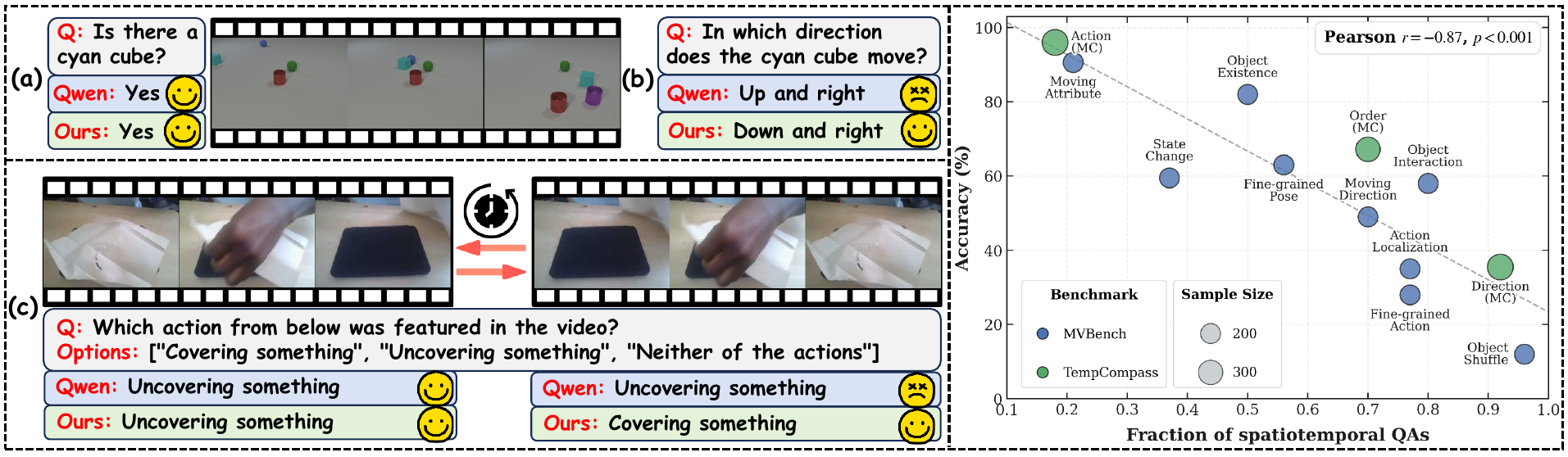}
  \caption{\textbf{High video-QA accuracy does not necessarily imply sensitivity to video dynamics.} \textbf{Left:} The model \textbf{(a)}~answers a static question correctly, but \textbf{(b)}~fails on a spatiotemporal question; \textbf{(c)}~it gives the same prediction to a video and its temporal reversal. \textbf{Right:} Across MVBench and TempCompass sub-tasks, accuracy drops as the fraction of spatiotemporal questions increases.}
  \label{fig:motivation}
  \vspace{-2mm}
\end{figure}

Video large language models (Video LLMs)~\citep{llavavideo,qwen3vl,internvl3,gemini3} have achieved strong results on video understanding benchmarks~\citep{videomme,mvbench}. Yet high answer accuracy does not necessarily imply that a model tracks spatiotemporal dynamics. Recent studies show that Video LLMs can often solve video questions using static shortcuts, such as single-frame cues or language priors, instead of following the underlying dynamics~\citep{lei2023revealing,zohar2025apollo,timeblind,krojer2025shortcut}. Figure~\ref{fig:motivation} illustrates this gap: Qwen3-VL~\citep{qwen3vl} can recognize a static property of a scene, but fails to track movement direction and gives the same prediction even after the video is temporally reversed. More generally, performance degrades on questions that require motion direction, event order, and other spatiotemporal information. These observations expose a capability that answer accuracy alone does not measure: a model should not only answer correctly, but also respond appropriately when the relevant video dynamics change.

This distinction is especially important in reinforcement-learning post-training. RL has become an effective way to improve reasoning in LLMs~\citep{grpo,dapo} and Video LLMs~\citep{videor1,VideoRFT,videochatr1}, but standard objectives primarily reward whether the final answer is correct. If a training question can be solved from a salient frame or a language prior, a shortcut policy can receive the same reward as a policy that genuinely tracks the video dynamics. Correctness-based RL therefore provides no explicit incentive for the prediction to respond when motion, temporal order, or other task-relevant evidence changes. A direct alternative would be to annotate which frames, motions, or events support each answer~\citep{yuan2025videorefer,shi2026videoloom}, but such process-level supervision is expensive and difficult to scale.

Our key observation is that \emph{controlled changes to the video provide a much simpler supervision signal}. Consider a question asking whether an object moves left or right. If an appropriate transformation changes the direction of motion, a model that uses this evidence should also change its answer. In contrast, if the question asks about the object's color or presence, the same transformation should leave the answer unchanged. This yields a simple principle: \emph{predictions should change when an intervention changes task-relevant evidence, and remain stable when it does not}. We refer to this property as \textbf{spatiotemporal sensitivity}. Importantly, this principle specifies only the expected \emph{relationship} between the predictions before and after the intervention. It therefore provides useful supervision even when the exact ground-truth answer to the transformed video is unavailable.

Based on this principle, we propose \textbf{Counterfactual Relational Policy Optimization (CRPO)}. For each video question, CRPO constructs a controlled counterfactual video using a task-appropriate spatial or temporal transformation, and generates rollouts from both the original and counterfactual inputs. The original branch is supervised by the ground-truth answer, while the counterfactual branch is trained according to how the answer should respond to the transformation. A \textbf{Counterfactual Relation Reward (CRR)} couples the two branches, encouraging the model to preserve its prediction when the queried property is unchanged and to alter it when task-relevant dynamics change. Since both branches update a shared policy, ground-truth supervision on the factual branch anchors answer correctness, while relational supervision constrains when predictions should change; Static examples additionally require predictions to remain stable when the queried property is preserved. Together, these signals discourage indiscriminate answer changes without requiring transformed-video labels.

CRPO also differs from prior perturbation-based video RL methods such as T-GRPO~\citep{videor1} and ArrowRL~\citep{xue2025seeing}. These methods use perturbed videos primarily to shape rewards for rollouts from the original video, whereas CRPO directly optimizes rollouts from both factual and counterfactual branches. We further identify an optimization issue in such reward-shaping approaches: when an auxiliary reward acts as a common rescaling within a GRPO rollout group, group-wise advantage normalization can cancel its effect on the policy gradient. By contrast, CRPO obtains rollout-level behavioral variation directly from the counterfactual branch. We analyze this effect in Section~\ref{sec:cancellation}.

We evaluate CRPO on paired spatiotemporal diagnostics and standard video benchmarks. Alongside existing benchmarks, we introduce \textbf{DyBench}, a paired diagnostic benchmark in which closely matched videos differ in their dynamics and therefore require different answers. Its strict pair accuracy requires both sides of a pair to be answered correctly, making fixed-answer shortcuts ineffective. Across Qwen3-VL-4B and Qwen3-VL-8B, CRPO consistently improves paired spatiotemporal evaluations while preserving performance on original videos and general video understanding. The gains also extend to Event Sequence examples constructed by segment reordering, a transformation never used during CRPO training. These results suggest that CRPO learns a broader sensitivity to video dynamics rather than simply fitting the training interventions.

\noindent Our contributions are threefold:
\begin{itemize}
    \item We formulate \textbf{spatiotemporal sensitivity} as a counterfactual relational learning problem: predictions should change when a controlled intervention alters task-relevant dynamics and remain stable when the queried property is preserved.

    \item We propose \textbf{CRPO}, a dual-branch RL framework that combines ground-truth supervision on factual videos with relational supervision on counterfactual videos under a shared policy, without requiring ground-truth answers for transformed videos or process-level spatiotemporal annotations.

    \item We introduce the paired \textbf{DyBench} diagnostic and conduct controlled and out-of-distribution evaluations showing that CRPO improves sensitivity to spatiotemporal changes while preserving general video understanding.
\end{itemize}

\section{Related Work}

\textbf{Video Large Language Models.}
The rapid progress of LLMs~\citep{gemini3,gpt5} has driven substantial advances in video understanding. Recent Video LLMs typically extend the image-language instruction-tuning paradigm~\citep{llava} to video by bridging video and text modalities through lightweight adapters or projectors, and by training on large-scale video-caption and video instruction-following data~\citep{llavaonevision,videollama3,llavavideo}. More recent open models~\citep{cho2025perceptionlm,molmo2,internvl3,qwen3vl} further strengthen multimodal reasoning, grounding, and long-context modeling through better data and training pipelines. Some work improves long-video understanding through memory, compression, or token reduction~\citep{song2024moviechat,videoxl,shen2024longvu}. Despite these advances, strong benchmark performance does not necessarily imply genuine spatiotemporal understanding, as many Video LLMs can still rely on static shortcuts such as single-frame cues or language priors~\citep{zohar2025apollo,krojer2025shortcut} rather than truly modeling how events unfold over time.

\textbf{RL for Video LLMs.}
Recent work has explored reinforcement learning for Video LLMs. GRPO~\citep{grpo} has become a common foundation for RL-based post-training, motivating studies on data selection, reward design, optimization stability, and credit assignment~\citep{videochatr1,VideoRFT,park2025deepvideo,feng2025onethinker,liu2026videoautor1,videoktr}. These works show that video RL depends critically on how reward signals are designed. Several recent methods further introduce video variations into RL. Video-R1~\citep{videor1} (T-GRPO) shuffles frames and rewards original-video rollouts when they outperform shuffled-video rollouts, while ArrowRL~\citep{xue2025seeing} uses temporal reversal to penalize original-video rollouts that mirror reversed-video responses. These perturbations provide useful contrastive signals, but the perturbed videos are used only for reward shaping. STRIVE~\citep{bahrami2026strive} groups rollouts jointly over textual outputs and spatiotemporal visual variants to improve exploration. CRPO follows the direction of using video variations, but treats the counterfactual video as a trainable branch: both original and counterfactual branches generate rollouts and contribute policy gradients. The Counterfactual Relation Reward then links the two branches by rewarding answer changes for dynamic questions and answer preservation for static questions, yielding a simple and targeted signal for spatiotemporal sensitivity.

\textbf{Video Understanding Benchmarks.} 
Standard benchmarks~\citep{videomme,mvbench,lvbench,zhao2025mmvu} cover broad video understanding abilities, but they do not explicitly diagnose shortcut-based success. Recent diagnostic benchmarks aim to close this gap. TempCompass~\citep{tempcompass} probes temporal perception with conflicting videos matched in static content. MHBench~\citep{mhbench} uses adversarial triplets to evaluate motion hallucination and test whether Video LLMs rely on static appearance instead of true motion perception. MVP~\citep{krojer2025shortcut} uses minimal-change video pairs with identical questions and opposite answers to reduce shortcut-based score inflation. TimeBlind~\citep{timeblind} is closest to our setting, using minimal video pairs and paired evaluation to isolate temporal understanding. Our proposed \emph{DyBench} follows this line, but frames paired evaluation through counterfactual video pairs. It focuses on three spatiotemporal tasks, namely reversible dynamics, moving direction, and event sequence, and uses \emph{pair accuracy} to distinguish shortcut success from genuine spatiotemporal understanding.

\section{Counterfactual Relational Policy Optimization}
\label{sec:method}

\begin{figure}[t]
  \centering
  \includegraphics[width=0.95\textwidth]{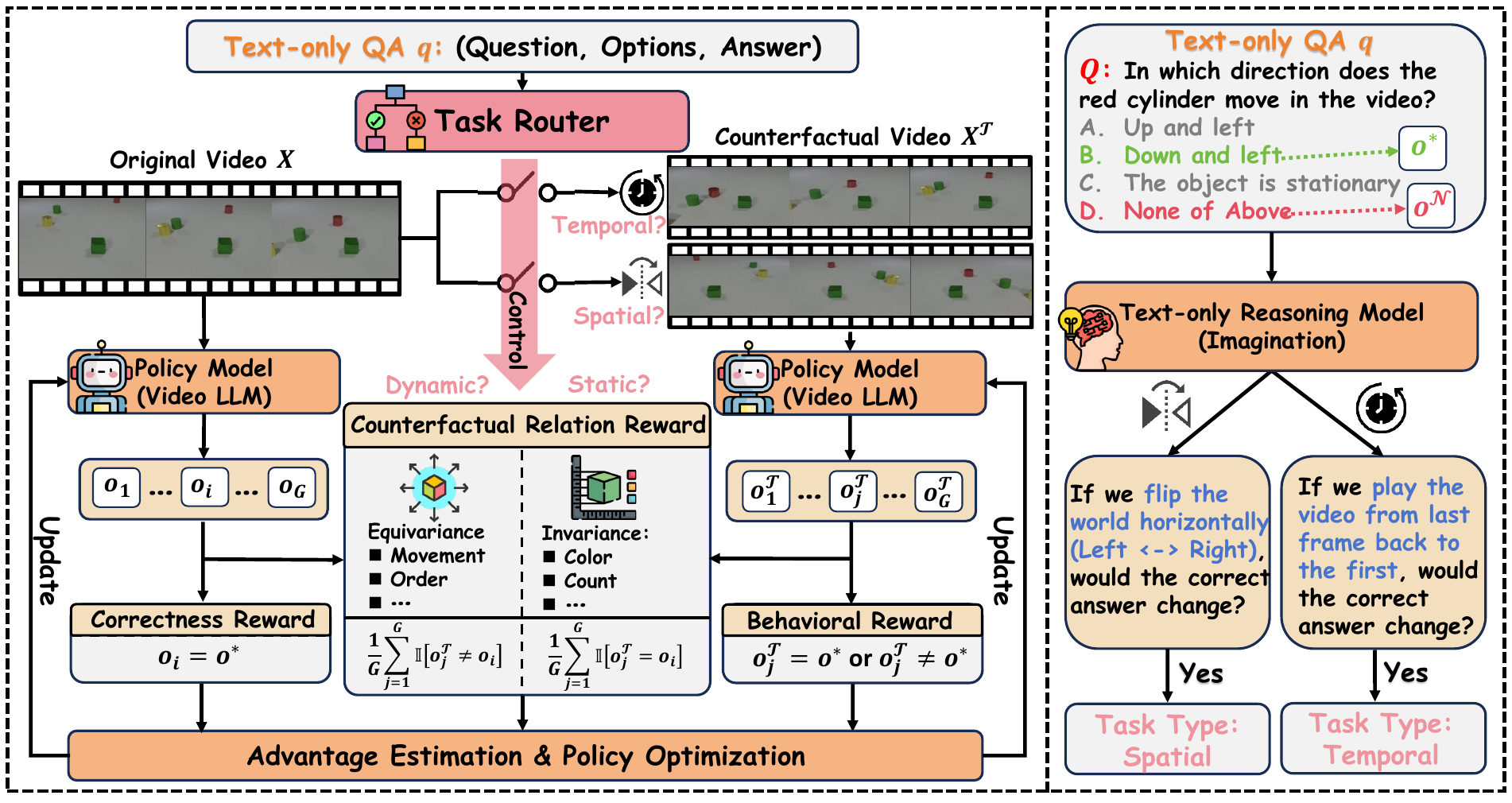}
  \caption{\textbf{Overview of CRPO.} \textbf{Left:} Given a video question, the Task Router selects a counterfactual transformation $\mathcal{T}$ (horizontal flip or temporal reversal). The original branch and the counterfactual branch each generate $G$ rollouts. These rollouts are scored by branch-specific correctness or behavioral rewards, format rewards, and the Counterfactual Relation Reward (CRR), and are then used for advantage estimation and policy optimization. \textbf{Right:} The Task Router is implemented as a text-only reasoning model that answers two hypothetical transformation questions to predict whether the correct answer would change under horizontal flip or temporal reversal.}
  \label{fig:crpo}
  \vspace{-3mm}
\end{figure}

Standard GRPO~\citep{grpo} optimizes whether a sampled response to a video question is correct, but does not require the prediction to respond when the underlying video dynamics change.
CRPO introduces this missing supervision by jointly training on a factual video and a controlled counterfactual version of it.
Given a training example $(X,Q,y)$, where $X$ is the video, $Q$ is the question, and $y$ is the ground-truth answer, we construct a counterfactual video $X^{\mathcal T}$ and generate rollout groups from both $(X,Q)$ and $(X^{\mathcal T},Q)$ using the same policy.
The factual branch provides supervision on \emph{what to answer}, while the counterfactual branch provides supervision on \emph{when the answer should change}.
Figure~\ref{fig:crpo} gives an overview of the framework.

\subsection{Task-Aware Counterfactual Construction}
\label{sec:counterfactual}

The effect of a visual transformation depends on what the question asks. For example, horizontally flipping a video can change the answer to a left--right motion question while leaving object color or identity unchanged. Similarly, temporal reversal can change event order or reversible actions such as opening and closing, while preserving static scene content.

As shown in the right panel of Figure~\ref{fig:crpo}, we determine this dependency offline using a text-only \textbf{Task Router}. Given only the question, answer options, and ground-truth answer, the router considers two hypothetical interventions: whether the correct answer would change if the video were horizontally flipped, and whether it would change if the video were played in reverse. The two decisions categorize the question as \textsc{Spatial}, \textsc{Temporal}, \textsc{Spatiotemporal}, or \textsc{Static}. Spatial questions use horizontal flipping, Temporal questions use temporal reversal, and Spatiotemporal questions randomly use one of the two transformations. For Static questions, either transformation can be used because the queried property should remain unchanged.
The router is executed once before RL training and introduces no additional inference cost during optimization. Its full prompt, classification statistics, and human validation are provided in Appendix~\ref{app:router}.

We refer to Spatial, Temporal, and Spatiotemporal questions collectively as \emph{dynamic}. For a dynamic question, the selected intervention changes task-relevant evidence and the answer is therefore expected to change. For a Static question, the queried property is preserved and the answer should remain unchanged. This gives CRPO a counterfactual supervision signal without requiring a new ground-truth label for $X^{\mathcal T}$.

\subsection{Counterfactual Relational Reward}
\label{sec:crr}

For each training example $(X,Q,y)$, CRPO samples $G$ factual rollouts
$\{o_i\}_{i=1}^{G}$ from the original video and $G$ counterfactual rollouts
$\{\tilde{o}_j\}_{j=1}^{G}$ from the transformed video $X^{\mathcal T}$.
Let $a_i$ and $\tilde a_j$ denote their extracted final answers.
The factual branch follows standard outcome supervision, with
$c_i=\mathbb{I}[a_i=y]$ indicating whether a rollout is correct.
The counterfactual branch instead evaluates whether the prediction responds to
the intervention as expected: for dynamic questions,
$b_j=\mathbb{I}[\tilde a_j\neq y]$, while for Static questions,
$b_j=\mathbb{I}[\tilde a_j=y]$.
We refer to this local counterfactual term as the behavioral reward,
$R_{\mathrm{behave}}(\tilde o_j)=b_j$.
We denote their group averages by
$\bar c=\frac{1}{G}\sum_i c_i$ and
$\bar b=\frac{1}{G}\sum_j b_j$.

CRPO couples factual correctness and counterfactual behavior through a symmetric
relational reward:
\begin{align}
R_i^{\mathrm{F}}
&=
c_i
+
\lambda_z c_i \bar b
+
R_{\mathrm{fmt}}(o_i),
\label{eq:factual_reward}
\\
R_j^{\mathrm{CF}}
&=
w_{\mathrm{aug}}
\left(
b_j
+
\lambda_z b_j \bar c
+
R_{\mathrm{fmt}}(\tilde{o}_j)
\right),
\label{eq:counterfactual_reward}
\end{align}
where $\lambda_z$ is $\lambda_d$ for dynamic questions and $\lambda_s$ for Static questions, and $w_{\mathrm{aug}}$ controls the overall contribution of the counterfactual branch. $R_{\mathrm{fmt}}$ is the standard format reward used to enforce the required response structure. The first term in each branch provides its local supervision, with factual correctness used for the original video and the expected behavioral response used for the counterfactual video, while the second term is the \textbf{Counterfactual Relation Reward (CRR)}. A correct factual rollout receives additional reward when the counterfactual
branch consistently responds to the intervention, and conversely, a counterfactual rollout is reinforced more strongly when the factual branch is reliably correct. CRR therefore encourages the two predictions to be jointly consistent with the underlying intervention rather than optimizing the two branches independently.

This decomposition reflects the complementary roles of the two supervision signals. Factual supervision anchors \emph{what the model should answer}, whereas counterfactual supervision constrains \emph{when that answer should change}. For dynamic questions, the relational signal specifies that the answer should change after the intervention, but does not determine the exact counterfactual answer, making it weaker than supervision from exact counterfactual labels. CRPO therefore does not optimize this signal in isolation. Because factual and counterfactual rollouts update the same policy, ground-truth supervision anchors factual correctness, while relational supervision encourages selective responses to task-relevant changes; Static examples constrain the policy to preserve predictions when the queried property remains unchanged. In this way, counterfactual relations serve as a behavioral constraint on a factually grounded policy, without requiring ground-truth labels for transformed videos.

A final practical issue arises for multiple-choice questions. After a spatial or temporal transformation, the correct counterfactual answer may no longer appear among the original answer choices. We therefore append a \emph{None of the above} option $o^{\mathcal{N}}$ (Null Option) during training. It is always incorrect for the factual video, where the original ground-truth answer remains available, but allows the counterfactual branch to represent cases in which the transformation moves the valid answer outside the original choice set. We analyze its selection behavior during training in
Appendix~\ref{app:null_frequency}.

\subsection{Dual-Branch Policy Optimization}
\label{sec:optimization}

Both factual and counterfactual rollouts directly optimize the same policy
$\pi_\theta$.
Following GRPO, rewards are centered within each rollout group.
CRPO centers the factual and counterfactual rewards separately, but normalizes
both groups using a shared standard deviation computed over the centered rewards
from the two branches.
This preserves the relative ordering of rollouts within each branch while
placing their advantages on a common optimization scale.
We denote the resulting advantages by
$\widehat A_k^{\mathrm F}$ and $\widehat A_k^{\mathrm{CF}}$.

Let $b\in\{\mathrm F,\mathrm{CF}\}$ index the factual and counterfactual
branches, and let $\rho_k^b$ denote the corresponding importance ratio between
the current and old policies.
CRPO optimizes
\begin{equation}
\mathcal{J}_{\mathrm{CRPO}}(\theta)
=
\sum_{b\in\{\mathrm{F},\mathrm{CF}\}}
\frac{1}{G}\sum_{k=1}^{G}
\min\!\left(
\rho_k^{b}\widehat A_k^{b},
\operatorname{clip}(\rho_k^{b},1-\epsilon,1+\epsilon)\widehat A_k^{b}
\right),
\label{eq:crpo_objective}
\end{equation}
with standard GRPO KL regularization applied to both branches.
Exact normalization details and the KL setting are provided in
Appendix~\ref{app:optimization}.

The key distinction from prior perturbation-based methods lies in how the
transformed video participates in optimization.
T-GRPO~\citep{videor1} and ArrowRL~\citep{xue2025seeing} primarily use
perturbed videos to modify the rewards of rollouts sampled from the original
input.
CRPO instead treats the counterfactual video as an explicit optimization branch:
its rollouts receive their own advantages and directly contribute policy
gradients to the shared model.
Thus, counterfactual information enters policy learning through the rollouts
themselves rather than only through an auxiliary reward on factual responses.
As we show next, this distinction also avoids an important limitation of
perturbation-based reward shaping under group-wise advantage normalization.

\subsection{Reward Cancellation under Group-Wise Advantage Normalization}
\label{sec:cancellation}

A key limitation of perturbation-based reward shaping is that its auxiliary
signal can be suppressed by the group-wise advantage normalization used in
GRPO.
This issue is particularly relevant to methods such as
T-GRPO~\citep{videor1} and ArrowRL~\citep{xue2025seeing}, where information
from the perturbed video is injected into the rewards of factual rollouts rather
than optimized through an independent rollout branch.

Consider the case in which the perturbation-derived signal assigns the same
bonus $C$ to every correct rollout in a group.
For binary correctness $c_i\in\{0,1\}$, with group mean $\bar c$ and standard
deviation $\sigma_c$, the normalized advantage becomes
\begin{equation}
\widehat A_i
=
\frac{(1+C)(c_i-\bar c)}{(1+C)\sigma_c+\epsilon}
\approx
\frac{c_i-\bar c}{\sigma_c}.
\label{eq:reward_cancellation}
\end{equation}
The common factor $(1+C)$ therefore cancels under normalization.
Although the auxiliary signal changes the numerical reward, it contributes
little or no additional relative advantage in this case, and hence provides
essentially no new policy-gradient signal.
This reveals an important limitation of injecting perturbation information only
through group-level reward shaping.

CRPO avoids this cancellation by optimizing counterfactual rollouts directly.
Because the counterfactual responses are independently sampled, their
behavioral rewards vary across rollouts according to whether each prediction
responds appropriately to the intervention.
These rollout-level differences survive group-wise normalization and produce
counterfactual advantages that directly update the shared policy.
Thus, counterfactual information enters CRPO through the policy gradient itself,
rather than only through a potentially normalized-away reward modifier.
The full derivation and additional analysis are provided in
Appendix~\ref{app:cancellation}.

\section{DyBench: A Paired Benchmark for Spatiotemporal Sensitivity}
\label{sec:dybench}

Existing video benchmarks are dominated by questions whose answers can often be
inferred from a single frame or a language prior~\citep{krojer2025shortcut,timeblind},
leaving open the question of whether a Video LLM truly tracks spatiotemporal
dynamics.
To diagnose this directly, we construct \textbf{DyBench}, a paired
counterfactual benchmark of 1{,}507 video pairs (3{,}014 videos) targeting
motion direction, event order, and the arrow of time, three aspects of video
understanding that are difficult to resolve from static cues alone.

\begin{figure}[t]
  \centering
  \includegraphics[width=0.95\textwidth]{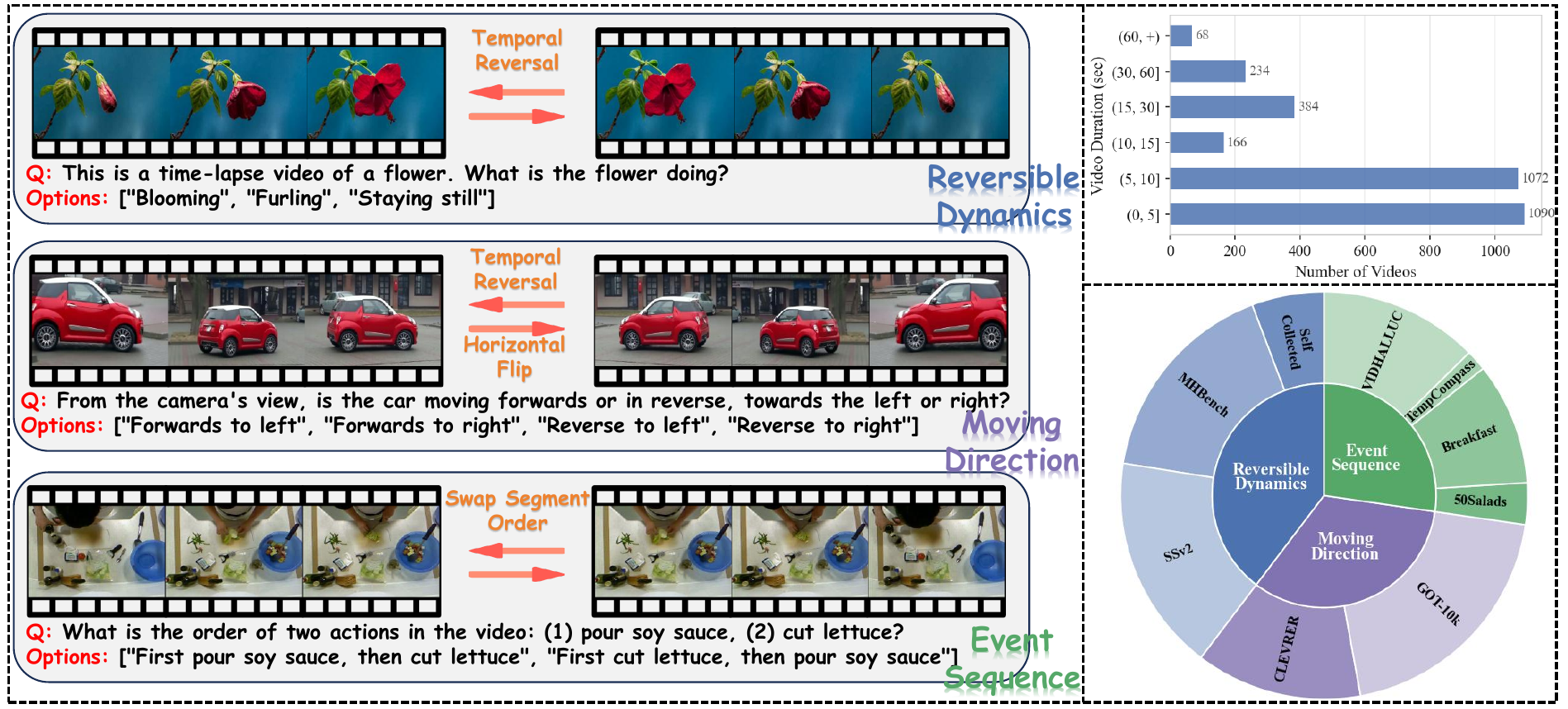}
  \caption{\textbf{Overview of DyBench.}
  DyBench contains 3{,}014 videos forming 1{,}507 counterfactual pairs across
  three sub-tasks: Reversible Dynamics, Moving Direction, and Event Sequence.
  The upper-right panel shows the distribution of video durations, while the
  lower-right panel shows the distribution of video sources.}
  \label{fig:dataset}
  \vspace{-3mm}
\end{figure}

\paragraph{Counterfactual pairs and pair accuracy.}
Each DyBench item is a \emph{counterfactual pair} $(v,v')$ rather than a
single video.
The two videos share the same underlying scene and objects, ask the same
multiple-choice question, and use the same semantic answer choices, while differing along a
controlled spatiotemporal axis that changes the correct answer.
We report \textbf{pair accuracy (P-Acc)}, which counts a pair as correct only
when the model answers both $v$ and $v'$ correctly.
By construction, a model that returns the same answer despite the changed
dynamics cannot succeed on the pair.
Compared with standard per-video accuracy (Acc), P-Acc therefore provides a
stricter measure of whether model predictions actually respond to the relevant
spatiotemporal change.

\paragraph{Three sub-tasks.}
DyBench covers three complementary forms of dynamic reasoning.
\textbf{(1) Reversible Dynamics} asks whether a process unfolds in one temporal
direction or the other, such as \emph{opening} versus \emph{closing} or
\emph{blooming} versus \emph{closing}, and constructs pairs through temporal
reversal.
\textbf{(2) Moving Direction} asks in which direction an object or actor moves,
with paired videos constructed through horizontal flipping and, when
appropriate, temporal reversal.
\textbf{(3) Event Sequence} asks about the order of two events and constructs
pairs by concatenating the same action segments in opposite orders.
As summarized in Figure~\ref{fig:dataset}, DyBench draws from diverse video
sources and covers a range of short-video durations.
Full construction details, source datasets, quality control, shortcut-isolation
analysis, and per-sub-task results are provided in
Appendix~\ref{sec:app_dybench}.

\section{Experiments}
\label{sec:exp}

\subsection{Experimental Setup}
\label{sec:setup}

\textbf{Benchmarks.}
We evaluate on five benchmarks.
DyBench (Section~\ref{sec:dybench}) and TimeBlind~\citep{timeblind} are paired
benchmarks for which we report both standard accuracy and a strict paired
metric: pair accuracy (P-Acc) for DyBench and instance accuracy (I-Acc) for
TimeBlind.
For DyBench, we additionally report \textbf{original accuracy (O-Acc)}, which
evaluates only the original, untransformed videos in each pair.
The other three benchmarks, TempCompass~\citep{tempcompass},
VideoMME~\citep{videomme}, and MVBench~\citep{mvbench}, use standard accuracy.
All evaluations are conducted with VLMEvalKit~\citep{vlmevalkit} using $32$
uniformly sampled frames.

\textbf{Baselines.}
We compare CRPO against two proprietary models, GPT-5.1~\citep{gpt5} and
Gemini-3.1-Pro~\citep{gemini3}, and several open-source Video LLMs, including
LLaVA-OneVision-7B~\citep{llavaonevision},
InternVL3-8B~\citep{internvl3},
Qwen2.5-VL-7B~\citep{qwen25vl}, and ArrowRL$^{*}$, the officially released
checkpoint of~\citet{xue2025seeing}.
We further compare CRPO with GRPO~\citep{grpo},
T-GRPO~\citep{videor1}, and ArrowRL~\citep{xue2025seeing} under controlled
training settings.
All RL post-training methods use the same backbone, training corpus, and shared
hyperparameters.
Further details are provided in Appendix~\ref{sec:app_baselines}.

\textbf{Training Details.}
We instantiate CRPO on Qwen3-VL-4B-Instruct and
Qwen3-VL-8B-Instruct~\citep{qwen3vl}, and follow the same training corpus and
RL hyperparameters as VideoAuto-R1~\citep{liu2026videoautor1}.
The corpus mixes text, image, and video QA; the dual-branch counterfactual
framework is applied only to the video subset, while text and image samples are
trained with standard single-branch GRPO.
The video subset contains $21{,}165$ samples and is processed with $G{=}8$
rollouts per branch, $32$ uniformly sampled frames per video, and a learning
rate of $1\times10^{-6}$.
We set $\lambda_d=\lambda_s=0.3$ and $w_{\mathrm{aug}}=0.5$.
The Task Router is run once offline using DeepSeek-R1.
Training is conducted on $32$ NVIDIA H20 GPUs.
Full data composition, reward definitions, hyperparameters, and Task Router
details are provided in Appendices~\ref{sec:app_data_reward},
\ref{sec:app_train_script}, and \ref{sec:app_router}.

\begin{table*}[t]
    \vspace{-4mm}
    \caption{\textbf{Main results on five video benchmarks.}
    O-Acc denotes accuracy on only the original, untransformed videos of
    DyBench, while P-Acc and I-Acc are the strict paired metrics of DyBench
    and TimeBlind, respectively.
    Values marked with $\uparrow$ denote absolute percentage-point improvements
    of CRPO over the corresponding Qwen3-VL base model.}
    \centering
    \setlength{\tabcolsep}{3.5pt}
    \arrayrulecolor{rulecolor}
    \resizebox{\textwidth}{!}{
    \begin{tabular}{l*{3}{c}*{2}{c}ccc}
        \thickrule
        \multicolumn{1}{c}{\multirow{3}{*}[-3pt]{\textbf{Models}}}
        & \multicolumn{6}{c}{
            \textcolor{groupclr}{\textbf{Spatiotemporal}}}
        & \multicolumn{2}{c}{
            \textcolor{groupclr}{\textbf{General}}} \\

        \cmidrule(lr){2-7} \cmidrule(lr){8-9}

        & \multicolumn{3}{c}{\textbf{DyBench}}
        & \multicolumn{2}{c}{\textbf{TimeBlind}}
        & \textbf{TempCompass}
        & \textbf{VideoMME}
        & \textbf{MVBench} \\

        \cmidrule(lr){2-4}
        \cmidrule(lr){5-6}
        \cmidrule(lr){7-7}
        \cmidrule(lr){8-8}
        \cmidrule(lr){9-9}

        & Acc & O-Acc & P-Acc
        & Acc & I-Acc
        & Acc & Acc & Acc \\
        \medrule

        \groupheader{9}{Proprietary Models} \\
        GPT-5.1~\citep{gpt5}
        & 63.7 & 76.7 & 44.9
        & 67.3 & 27.0
        & 76.4 & 72.9 & 62.9 \\

        Gemini-3.1-Pro~\citep{gemini3}
        & 82.2 & 88.3 & 71.7
        & 77.2 & 45.5
        & 74.5 & 74.3 & 73.6 \\
        \medrule

        \groupheader{9}{Open-Source Models} \\
        LLaVA-OV-7B~\citep{llavaonevision}
        & 48.4 & 59.1 & 19.6
        & 56.5 & 7.8
        & 60.5 & 57.6 & 54.7 \\

        InternVL3-8B~\citep{internvl3}
        & 69.5 & 79.0 & 50.2
        & 63.0 & 18.3
        & 70.5 & 65.6 & 74.3 \\

        Qwen2.5-VL-7B~\citep{qwen25vl}
        & 64.5 & 76.0 & 44.1
        & 65.0 & 22.5
        & 70.9 & 60.3 & 67.1 \\

        ArrowRL$^{*}$~\citep{xue2025seeing}
        & 67.5 & 78.8 & 50.1
        & 65.0 & 20.2
        & 71.7 & 60.1 & 66.2 \\
        \medrule

        \groupheader{9}{RL Post-Training of Qwen3-VL-4B} \\
        Qwen3-VL-4B (base)~\citep{qwen3vl}
        & 65.1 & 76.2 & 45.4
        & 66.2 & 26.5
        & 71.6 & 62.2 & 67.5 \\

        \quad + GRPO~\citep{grpo}
        & 66.6 & 78.0 & 48.2
        & 67.8 & 28.0
        & 72.2 & 62.1 & 68.7 \\

        \quad + T-GRPO~\citep{videor1}
        & 68.6 & 78.5 & 49.8
        & 67.9 & 29.0
        & 73.5 & \textbf{63.4} & \textbf{70.3} \\

        \quad + ArrowRL~\citep{xue2025seeing}
        & 66.7 & 77.6 & 47.5
        & 67.5 & 26.8
        & 71.6 & 62.2 & 69.9 \\

        \rowcolor{ourshade}
        \quad + \ours{CRPO (Ours)}
        & \textbf{70.3}\up{5.2}
        & \textbf{79.5}\up{3.3}
        & \textbf{54.8}\up{9.4}
        & \textbf{69.8}\up{3.6}
        & \textbf{31.7}\up{5.2}
        & \textbf{74.2}\up{2.6}
        & 63.0\up{0.8}
        & 68.6\up{1.1} \\
        \medrule

        \groupheader{9}{RL Post-Training of Qwen3-VL-8B} \\
        Qwen3-VL-8B (base)~\citep{qwen3vl}
        & 68.2 & 78.9 & 50.4
        & 67.8 & 27.8
        & 75.1 & 64.9 & 68.7 \\

        \quad + GRPO~\citep{grpo}
        & 69.4 & 79.9 & 52.4
        & 69.8 & 30.5
        & 75.5 & 64.5 & 69.1 \\

        \quad + T-GRPO~\citep{videor1}
        & 70.9 & 80.8 & 54.5
        & 70.3 & 32.5
        & 75.4 & 65.0 & 69.5 \\

        \quad + ArrowRL~\citep{xue2025seeing}
        & 69.6 & 80.3 & 52.2
        & 69.3 & 29.3
        & 74.9 & 65.1 & 69.7 \\

        \rowcolor{ourshade}
        \quad + \ours{CRPO (Ours)}
        & \textbf{72.5}\up{4.3}
        & \textbf{81.6}\up{2.7}
        & \textbf{58.1}\up{7.7}
        & \textbf{71.7}\up{3.9}
        & \textbf{36.0}\up{8.2}
        & \textbf{77.4}\up{2.3}
        & \textbf{65.6}\up{0.7}
        & \textbf{69.7}\up{1.0} \\
        \thickrule
    \end{tabular}}
    \label{tab:main_results}
\end{table*}

\subsection{Main Results}
\label{sec:main}

Table~\ref{tab:main_results} summarizes the results across the five benchmarks.
\textbf{CRPO consistently improves spatiotemporal sensitivity.}
On Qwen3-VL-4B, CRPO improves DyBench P-Acc by $+9.4$ and TimeBlind I-Acc by
$+5.2$ over the base model, and further outperforms T-GRPO by $+5.0$ and
$+2.7$ points, respectively.
The same trend holds on Qwen3-VL-8B, where CRPO exceeds T-GRPO by $+3.6$
P-Acc on DyBench and $+3.5$ I-Acc on TimeBlind.
These gains are larger on the strict paired metrics than on standard accuracy,
indicating that CRPO particularly improves the model's response to changes in
video dynamics.
Importantly, this does not come from specializing to transformed test inputs:
on the original, untransformed DyBench videos, CRPO improves O-Acc from $76.2$
to $79.5$ for 4B and from $78.9$ to $81.6$ for 8B.
The improvement also extends to Event Sequence, which uses segment reordering
rather than the transformations seen during training; full per-sub-task results
are provided in Appendix~\ref{sec:app_subtask}, with paired qualitative examples
in Appendix~\ref{sec:app_examples}. \textbf{CRPO preserves general video understanding and scales to the larger
backbone.}
CRPO improves over the corresponding base model on TempCompass, VideoMME, and
MVBench for both 4B and 8B.
On the 4B backbone, T-GRPO remains stronger on VideoMME and MVBench, while CRPO
achieves substantially better paired spatiotemporal performance.
On the 8B backbone, CRPO obtains the best RL result on VideoMME and ties the
best result on MVBench, while improving DyBench P-Acc and TimeBlind I-Acc by
$+7.7$ and $+8.2$ over the base model.
Overall, CRPO improves sensitivity to spatiotemporal changes while maintaining
the model's general video-understanding capability.

\subsection{More Analysis}
\label{sec:analysis}

\begin{table*}[t]
    \vspace{-4mm}
    \caption{\textbf{Ablation on Qwen3-VL-4B.}
    Top: controls for rollout computation and data exposure together with
    controlled RL baselines.
    Bottom: removing individual components of CRPO. The ``w/o Counterfactual Branch'' setting removes direct policy updates from the counterfactual branch while retaining its rollouts for the factual-branch relational reward.}
    \centering
    \setlength{\tabcolsep}{3.5pt}
    \arrayrulecolor{rulecolor}
    \resizebox{0.97\textwidth}{!}{
    \begin{tabular}{l*{3}{c}*{2}{c}ccc}
        \thickrule
        \multicolumn{1}{c}{\multirow{2}{*}[-3pt]{\textbf{RL Method}}}
        & \multicolumn{3}{c}{\textbf{DyBench}}
        & \multicolumn{2}{c}{\textbf{TimeBlind}}
        & \textbf{TempCompass}
        & \textbf{VideoMME}
        & \textbf{MVBench} \\
        \cmidrule(lr){2-4}
        \cmidrule(lr){5-6}
        \cmidrule(lr){7-7}
        \cmidrule(lr){8-8}
        \cmidrule(lr){9-9}
        & Acc & O-Acc & P-Acc
        & Acc & I-Acc
        & Acc & Acc & Acc \\
        \medrule

        \groupheader{9}{Baseline} \\
        Qwen3-VL-4B (no RL)
        & 65.1 & 76.2 & 45.4
        & 66.2 & 26.5
        & 71.6 & 62.2 & 67.5 \\
        \medrule

        \groupheader{9}{Data / Compute Controls and RL Baselines} \\
        GRPO
        & 66.6 & 78.0 & 48.2
        & 67.8 & 28.0
        & 72.2 & 62.1 & 68.7 \\

        GRPO (G$\times$2)
        & 68.5 & 78.6 & 50.2
        & 68.5 & 29.0
        & 74.0 & 62.3 & 70.2 \\

        GRPO (Data$\times$2)
        & 66.5 & 78.0 & 48.4
        & 67.4 & 27.3
        & 73.0 & 62.0 & 69.8 \\

        T-GRPO
        & 68.6 & 78.5 & 49.8
        & 67.9 & 29.0
        & 73.5 & \textbf{63.4} & \textbf{70.3} \\

        ArrowRL
        & 66.7 & 77.6 & 47.5
        & 67.5 & 26.8
        & 71.6 & 62.2 & 69.9 \\
        \medrule

        \rowcolor{ourshade}
        \ours{CRPO (Ours)}
        & \textbf{70.3} & \textbf{79.5} & \textbf{54.8}
        & \textbf{69.8} & \textbf{31.7}
        & \textbf{74.2} & 63.0 & 68.6 \\
        \medrule

        \groupheader{9}{Component Ablations} \\
        w/o Spatial Flip
        & 69.5 & 79.0 & 53.2
        & 69.2 & 31.4
        & 74.0 & 62.8 & 69.0 \\

        w/o Temporal Reversal
        & 68.9 & 78.8 & 50.7
        & 68.3 & 29.6
        & 73.1 & 62.3 & 68.8 \\

        w/o CRR
        & 69.3 & 78.9 & 51.2
        & 68.8 & 30.5
        & 73.1 & 63.0 & 67.5 \\


        w/o Counterfactual Branch
        & 67.5 & 78.5 & 48.8
        & 68.0 & 29.4
        & 72.5 & 62.8 & 68.2 \\

        w/o Null Option
        & 65.9 & 76.6 & 50.1
        & 68.1 & 28.5
        & 72.2 & 62.1 & 63.7 \\
        \thickrule
    \end{tabular}}
    \label{tab:ablation}
\end{table*}

\textbf{Controlled ablations.}
Table~\ref{tab:ablation} shows that CRPO's gains cannot be explained by
additional data or rollout computation alone.
GRPO (G$\times$2) remains $4.6$ points below CRPO on DyBench P-Acc and $2.7$
points below on TimeBlind I-Acc, while GRPO (Data$\times$2) provides little
improvement over standard GRPO.
Removing the counterfactual relation reward (CRR) degrades paired performance, and removing the counterfactual branch causes the largest drop in
DyBench P-Acc, from $54.8$ to $48.8$.
Both counterfactual transformations contribute, with temporal reversal having
a larger effect than spatial flipping, while removing the Null Option also
causes substantial degradation. Additional diagnostics show that CRPO remains robust to moderate Task Router
errors and that the Null Option is used selectively rather than collapsing
across branches; see Appendices~\ref{app:router_robustness}
and~\ref{app:null_frequency}.

\begin{figure}[t]
  \centering
  \includegraphics[width=0.97\textwidth]{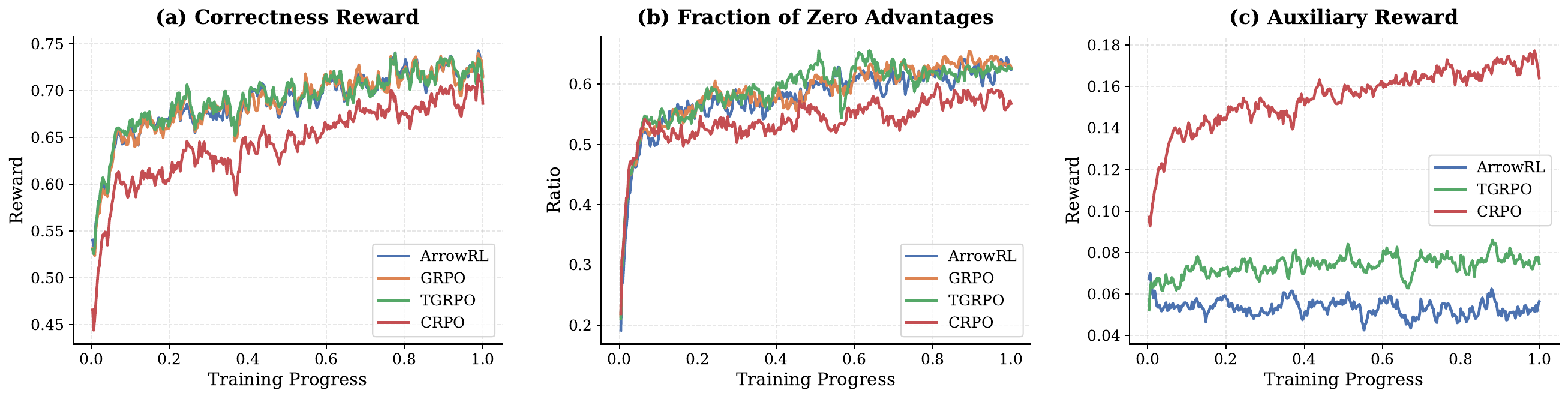}
  \caption{\textbf{Training dynamics of CRPO vs.\ RL baselines.}
  (a)~factual correctness reward, (b)~fraction of zero-advantage rollout groups,
  and (c)~auxiliary or relational reward.}
  \label{fig:reward}
  \vspace{-3mm}
\end{figure}

\textbf{Training-curve analysis.}
Figure~\ref{fig:reward} compares the training dynamics of CRPO, GRPO, T-GRPO,
and ArrowRL. Three observations stand out. \textbf{(a) Correctness reward.}
GRPO, T-GRPO, and ArrowRL rapidly increase factual correctness, reaching
approximately $0.70$ within the first $20\%$ of training, whereas CRPO rises
more gradually before converging to a comparable final level, indicating that
its gains do not come from optimizing factual correctness more aggressively.
\textbf{(b) Fraction of zero advantages.} CRPO consistently exhibits fewer
rollout groups with all-equal rewards and hence zero advantages, as the
independently sampled counterfactual responses introduce additional
rollout-level reward variation even when the factual branch is already highly
accurate, providing a denser optimization signal throughout training.
\textbf{(c) Auxiliary reward.} The perturbation-derived rewards of T-GRPO and
ArrowRL remain low and relatively flat, consistent with the normalization
effect analyzed in Section~\ref{sec:cancellation}, whereas CRPO's relational
reward increases steadily, indicating that the policy progressively learns to
satisfy the desired relation between factual and counterfactual predictions.
Together, these dynamics show that directly optimizing counterfactual rollouts
provides a more persistent learning signal than correctness-only RL or
perturbation-based reward shaping.

\section{Conclusion}
\label{sec:conclusion}

We present \textbf{CRPO}, a dual-branch reinforcement-learning framework for
training Video LLMs to be sensitive to task-relevant spatiotemporal changes.
CRPO combines factual and counterfactual supervision under a shared policy:
factual supervision anchors what the model should answer, while counterfactual
supervision constrains when that answer should change.
By directly optimizing counterfactual rollouts, CRPO provides a direct
spatiotemporal learning signal while avoiding the normalization-induced
limitation of perturbation-based reward shaping.
We also introduce \textbf{DyBench}, a paired benchmark with strict pair accuracy
for measuring this behavior.
Experiments show that CRPO improves spatiotemporal sensitivity and video understanding, with gains extending to an unseen transformation.
Together, these results show that controlled counterfactual relations provide a simple but practical way to ground Video LLM predictions in visual dynamics.

\section*{AI Use Statement}

In this work, we used generative AI tools as a component of the proposed method
and as a general-purpose writing aid.
Specifically, DeepSeek-R1 is used offline as the Task Router
(Section~\ref{sec:counterfactual}, Appendix~\ref{sec:app_router}) to classify
training questions according to their expected response to spatial and temporal
transformations.
The reliability of this component is evaluated through a human audit reported
in Appendix~\ref{sec:app_router}.
We additionally used large language models to polish the wording of the
manuscript and to assist with routine code and plotting scripts.
All AI-assisted content was reviewed by the authors, who take full
responsibility for the final manuscript and released artifacts.

\section*{Ethics Statement}

This work does not involve human subjects, personally identifying information,
or sensitive attributes.
DyBench is constructed from existing research video datasets
(Something-Something-V2, MHBench, GOT-10k, CLEVRER, Breakfast, 50Salads,
VIDHALLUC, and TempCompass) together with web-sourced clips. We use these
materials for research evaluation and follow the access and usage terms of the
source datasets.
The counterfactual transformations used in DyBench, including horizontal
flipping, temporal reversal, and segment reordering, are geometric or temporal
operations that do not synthesize new depicted individuals.
Improving the spatiotemporal sensitivity of Video LLMs has the same broad
dual-use considerations as video understanding more generally.
Limitations and intended applications are discussed in the appendix.
We declare no conflicts of interest.

\section*{Reproducibility Statement}

The CRPO objective, Counterfactual Relation Reward, counterfactual construction,
and dual-branch optimization are specified in
Section~\ref{sec:method}.
The training-data composition and standard reward definitions are provided in
Appendix~\ref{sec:app_data_reward}, the full training configuration in
Appendix~\ref{sec:app_train}, and the Task Router prompt and validation in
Appendix~\ref{sec:app_router}. Exact normalization and Null-Option details are
provided in Appendix~\ref{sec:app_method}.
Baseline RL algorithms are described in
Appendix~\ref{sec:app_baselines}; all controlled RL comparisons use the same
backbone, training corpus, and optimization settings.
The construction protocol, source datasets, and quality-control procedure for
DyBench are documented in Appendix~\ref{sec:app_dybench}.
All main evaluations use VLMEvalKit with $32$ uniformly sampled frames, as
described in Section~\ref{sec:setup}.
Code, DyBench, and model checkpoints will be released.

\bibliographystyle{iclr2027_conference}
\bibliography{iclr2027_conference}


\newpage
\appendix

\section{DyBench: Construction, Comparison, and Detailed Analysis}
\label{sec:app_dybench}

\subsection{Design principle and held-out evaluation}
The design of DyBench follows the recent observation that Video LLMs frequently solve dynamic questions through static shortcuts~\citep{krojer2025shortcut,timeblind}. Following the temporal-minimal-pair protocol popularized by TimeBlind~\citep{timeblind}, every DyBench item is built as a pair $(v, v')$ of videos that share the same scene and objects but differ along a single dynamic axis. The matching question $q$ has different correct answers for $v$ and $v'$, so a model that exploits a single keyframe or a language prior cannot simultaneously satisfy both. Pair accuracy then reports the fraction of pairs answered correctly on \emph{both} sides, providing a much stricter measure of dynamic understanding than per-question accuracy.

DyBench is also a strict out-of-distribution evaluation. None of its source datasets (something-something-v2~\citep{goyal2017something}, MHBench~\citep{mhbench}, GOT-10k~\citep{huang2019got}, the Breakfast cooking dataset~\citep{breakfast}, etc.) appear in the training pool used by any of the RL methods we benchmark. Our training pool consists of Video-R1, TVBench, STI-Bench, and MMR-VBench (Appendix~\ref{sec:app_train}). This source-level separation makes DyBench a held-out evaluation and reduces direct training--evaluation overlap as an explanation for the reported gains.

\subsection{Sub-task construction}

\paragraph{Sub-task 1: Reversible Dynamics.}
We collect actions whose semantics flip under temporal reversal. The action vocabulary covers humans, animals, and plants:
\begin{itemize}
  \item \emph{Human actions} sourced from MHBench~\citep{mhbench}, something-something-v2~\citep{goyal2017something}, and web-crawled clips: opening$\leftrightarrow$closing, picking up$\leftrightarrow$putting down, lifting$\leftrightarrow$throwing, hanging$\leftrightarrow$taking down, raising$\leftrightarrow$lowering, folding$\leftrightarrow$unfolding, pushing$\leftrightarrow$pulling, plugging in$\leftrightarrow$unplugging, zipping$\leftrightarrow$unzipping, rolling$\leftrightarrow$unrolling, tying$\leftrightarrow$untying, fastening$\leftrightarrow$unfastening, turning on$\leftrightarrow$turning off, getting in$\leftrightarrow$getting out, putting on$\leftrightarrow$taking off, inserting$\leftrightarrow$removing, stacking$\leftrightarrow$unstacking, sitting/lying/kneeling/squatting down$\leftrightarrow$standing up, inflating$\leftrightarrow$deflating, tightening$\leftrightarrow$loosening, assembling$\leftrightarrow$disassembling, leaning forward$\leftrightarrow$leaning back, bowing$\leftrightarrow$straightening up, drawing curtain open$\leftrightarrow$closed, walking forward$\leftrightarrow$backward, opening eyes$\leftrightarrow$closing eyes.
  \item \emph{Animal/plant actions} sourced from web-crawled clips: standing up$\leftrightarrow$lying down, jumping up$\leftrightarrow$landing, jumping into water$\leftrightarrow$leaping out, spreading wings$\leftrightarrow$folding wings, ears up$\leftrightarrow$flattened, arching back$\leftrightarrow$relaxing, retracting into shell$\leftrightarrow$extending out, stretching neck out$\leftrightarrow$retracting, opening mouth$\leftrightarrow$closing, extending tongue$\leftrightarrow$retracting, elephant raising trunk$\leftrightarrow$lowering trunk, raising head$\leftrightarrow$lowering head, flower blooming$\leftrightarrow$closing, leaf unfolding$\leftrightarrow$folding.
\end{itemize}
For each source clip we generate $v$ (forward play) and $v'$ (reverse play). The question asks ``Which of the following best describes the action in the video?'' with options drawn from the forward action, the reversed action, and a distractor. The answer flips between $v$ and $v'$.

\paragraph{Sub-task 2: Moving Direction.}
We use object-tracking videos from GOT-10k~\citep{huang2019got} and motion-direction questions from CLEVRER~\citep{yi2019clevrer}. For each source clip we generate $v$ (original) and $v'$ (horizontally flipped, optionally also temporally reversed). The question asks ``In what direction is the [object] moving?'' with the four standard direction options. For tracking videos with non-axial motion (e.g., diagonals), the option set is augmented with the four diagonals.

\paragraph{Sub-task 3: Event Sequence.}
We assemble pairs of two-event composite videos from four sources, including Breakfast~\citep{breakfast}, 50Salads~\citep{50salads}, VIDHALLUC~\citep{vidhalluc} and TempCompass~\citep{tempcompass}. For Breakfast and 50Salads, we follow the original action annotations and sample two non-adjacent action segments per video (each $\le 30$s). The two segments are concatenated to form $v$ and concatenated in the reverse order to form $v'$. For VIDHALLUC, we use cached DINO features to find the splice point, keeping only those clips with exactly one detected splice. Each clip is then re-spliced in reverse order to form $v'$.

\paragraph{Quality control.}
Every pair undergoes manual verification of (i)~content consistency between $v$ and $v'$, (ii)~counterfactual validity, so that no unintended task-relevant cue changes beyond the designed intervention, and (iii)~unambiguous correctness of the answer for both videos. Pairs that fail any check are discarded.

\subsection{Comparison with existing video benchmarks}
\label{sec:app_bench_compare}
Table~\ref{tab:app_bench_compare} compares DyBench with widely used video benchmarks along four dimensions: scale, video length, domain focus, and whether the benchmark uses adversarial pair construction. DyBench is most closely related to TimeBlind~\citep{timeblind} in spirit, since both use minimal pair construction and paired evaluation, but DyBench focuses specifically on three motion-aware sub-tasks (reversible dynamics, moving direction, and event sequence) and reports pair accuracy as the primary metric. In addition, the DyBench dataset is much larger.

\begin{table}[htbp]
\centering
\caption{DyBench vs.\ existing video benchmarks. ``Adversarial'' means each item is paired with a counter-example designed to flip the answer.}
\label{tab:app_bench_compare}
\small
\setlength{\tabcolsep}{8pt}
\renewcommand{\arraystretch}{1.18}
\arrayrulecolor{rulecolor}
\begin{tabular}{l c c c c}
\thickrule
\textbf{Benchmark} & \textbf{\#Videos / \#QAs} & \textbf{Length} & \textbf{Domain} & \textbf{Adversarial} \\
\medrule
MVBench~\citep{mvbench}              & 4{,}000 / 4{,}000   & short        & general       & \textcolor{red!70!black}{No}  \\
VideoMME~\citep{videomme}            & 900 / 2{,}700       & short, long  & general       & \textcolor{red!70!black}{No}  \\
LVBench~\citep{lvbench}              & 103 / 1{,}549       & long         & general       & \textcolor{red!70!black}{No}  \\
TempCompass~\citep{tempcompass}      & 410 / 1{,}580       & short        & dynamic       & \textcolor{green!50!black}{Yes} \\
TemporalBench~\citep{temporalbench}  & 2{,}000 / 10{,}000  & short, long  & dynamic       & \textcolor{red!70!black}{No}  \\
MotionBench~\citep{hong2025motionbench}                      & 5{,}385 / 8{,}052   & short        & human action  & \textcolor{red!70!black}{No}  \\
MHBench~\citep{mhbench}              & 1{,}200 / 3{,}600   & short        & human action  & \textcolor{green!50!black}{Yes} \\
TimeBlind~\citep{timeblind}          & 1{,}200 / 2{,}400   & short        & dynamic       & \textcolor{green!50!black}{Yes} \\
\medrule
\rowcolor{ourshade}
\ours{DyBench (Ours)}               & \textbf{3{,}014 / 3{,}014} & short & dynamic      & \textcolor{green!50!black}{\textbf{Yes}} \\
\thickrule
\end{tabular}
\end{table}

\subsection{Shortcut-isolation analysis}
\label{sec:app_shortcut}

A central design goal of DyBench is to make static shortcuts ineffective under paired evaluation. We verify this by re-evaluating models under three handicapped input settings, in the spirit of the shortcut analyses of TimeBlind~\citep{timeblind} and MVP~\citep{krojer2025shortcut}.
\begin{itemize}\setlength\itemsep{0pt}
  \item \textbf{Single Frame.} The model receives only a single, randomly sampled frame from each video.
  \item \textbf{Shuffled Frames.} The model receives all 32 frames but in a uniformly random order, destroying temporal structure while preserving every static cue.
  \item \textbf{Text Only.} The model receives the question and options without any visual input.
\end{itemize}
A benchmark that is genuinely diagnostic of spatiotemporal understanding should drive accuracy to near-chance under all three settings. Table~\ref{tab:app_shortcut} reports the results. Because DyBench mixes 2-, 3-, and 4-way questions, the random-chance overall is $35.1\%$ for Acc and $13.3\%$ for P-Acc; per sub-task, the chance values are $33.3\%/11.1\%$ for Reversible Dynamics (3-way), $25.0\%/6.3\%$ for Moving Direction (4-way), and $50.0\%/25.0\%$ for Event Sequence (binary).

\begin{table}[htbp]
  \centering
  \caption{\textbf{Shortcut-isolation analysis on DyBench.} ``Full Video'' is the standard 32-frame setting reproduced from Table~\ref{tab:main_results}; the other three columns are handicapped inputs that destroy temporal information.}
  \label{tab:app_shortcut}
  \small
  \setlength{\tabcolsep}{6pt}
  \renewcommand{\arraystretch}{1.18}
  \arrayrulecolor{rulecolor}
  \begin{tabular}{l *{4}{c c}}
    \thickrule
    \multirow{2}{*}{\textbf{Model}}
      & \multicolumn{2}{c}{\textbf{Full Video}}
      & \multicolumn{2}{c}{\textbf{Single Frame}}
      & \multicolumn{2}{c}{\textbf{Shuffled Frames}}
      & \multicolumn{2}{c}{\textbf{Text Only}} \\
      \cmidrule(lr){2-3}\cmidrule(lr){4-5}\cmidrule(lr){6-7}\cmidrule(lr){8-9}
      & Acc & P-Acc & Acc & P-Acc & Acc & P-Acc & Acc & P-Acc \\
    \medrule
    GPT-5.1                       & 63.7 & 44.9 & 39.9   & 5.1   & 30.5   & 3.0   & 32.8   & 2.6   \\
    Gemini-3.1-Pro                & 82.2 & 71.7 & 43.3  & 6.8  & 41.6   & 7.7   & 42.8  & 6.1   \\
    Qwen3-VL-8B                   & 68.2 & 50.4 & 35.1   & 7.2   & 42.3   & 10.2   & 27.7   & 0.4   \\
    \rowcolor{ourshade}
    \quad + \ours{CRPO (Ours)}    & 72.5 & 58.1 &  37.9  & 7.4   & 44.3   & 13.2   & 36.8   & 1.1   \\
    \medrule
    Random chance                 & 35.1 & 13.3 & 35.1 & 13.3 & 35.1 & 13.3 & 35.1 & 13.3 \\
    \thickrule
  \end{tabular}
\end{table}

We find that \textbf{under every handicapped input, all models collapse to or below chance on the strict pair metric}. CRPO drops from 58.1 P-Acc on the full video to 7.4 (Single Frame), 13.2 (Shuffled Frames), and 1.1 (Text Only), well below the 13.3 chance level for two of the three settings. The same pattern holds for GPT-5.1 (P-Acc collapses from 44.9 to $\le 5.1$), Gemini-3.1-Pro and the Qwen3-VL-8B baseline. This validates DyBench as a benchmark that genuinely requires temporal information, since shuffling frames, removing all but one frame, or removing the video entirely all destroy the signal needed to solve it. The Text-Only P-Acc values are small but not exactly zero because some construction pipelines preserve the same semantic answer choices while changing their presentation order across the two sides of a pair, so the textual prompts are not always byte-identical.

\subsection{Per-sub-task breakdown}
\label{sec:app_subtask}

Table~\ref{tab:app_subtask} decomposes DyBench accuracy into the three sub-tasks introduced in Sec.~\ref{sec:dybench}.

\begin{table}[htbp]
  \centering
  \caption{\textbf{Per-sub-task DyBench results.} Both per-question accuracy (Acc) and pair accuracy (P-Acc) are reported on each sub-task.}
  \label{tab:app_subtask}
  \small
  \setlength{\tabcolsep}{6pt}
  \renewcommand{\arraystretch}{1.18}
  \arrayrulecolor{rulecolor}
  \resizebox{\textwidth}{!}{
  \begin{tabular}{l *{2}{c} *{2}{c} *{2}{c} *{2}{c}}
    \thickrule
    \multirow{2}{*}{\textbf{Model}}
      & \multicolumn{2}{c}{\textbf{Reversible Dynamics}}
      & \multicolumn{2}{c}{\textbf{Moving Direction}}
      & \multicolumn{2}{c}{\textbf{Event Sequence}}
      & \multicolumn{2}{c}{\textbf{Overall}} \\
      \cmidrule(lr){2-3}\cmidrule(lr){4-5}\cmidrule(lr){6-7}\cmidrule(lr){8-9}
      & Acc & P-Acc & Acc & P-Acc & Acc & P-Acc & Acc & P-Acc \\
    \medrule
    GPT-5.1                       & 80.7 & 66.0 & 31.2 & 9.6 & 78.5 & 57.0 & 63.7 & 44.9 \\
    Gemini-3.1-Pro                & 96.1 & 92.3 & 57.7 & 36.8 & 91.9 & 84.3 & 82.2 & 71.7 \\
    \medrule
    Qwen3-VL-4B (base)            & 78.9 & 64.3 & 40.1 & 18.0 & 75.4 & 51.1 & 65.1 & 45.4 \\
    \quad + GRPO                  & 82.0 & 67.3 & 39.0 & 19.0 & 77.9 & 56.0 & 66.6 & 48.2 \\
    \quad + T-GRPO                & 82.0 & 68.3 & \textbf{44.7} & \textbf{21.8} & 78.3 & 57.0 & 68.6 & 49.8 \\
    \quad + ArrowRL               & 80.8 & 66.4 & 40.1 & 17.2 & 78.5 & 57.0 & 66.7 & 47.5 \\
    \rowcolor{ourshade}
    \quad + \ours{CRPO (Ours)}    & \textbf{86.8} & \textbf{77.7} & 41.8 & 21.0 & \textbf{81.1} & \textbf{62.7} & \textbf{70.3} & \textbf{54.8} \\
    \medrule
    Qwen3-VL-8B (base)            & 83.3 & 70.8 & 40.8 & 18.2 & 79.5 & 59.7 & 68.2 & 50.4 \\
    \quad + GRPO                  & 84.0 & 71.8 & 41.4 & 17.6 & 82.4 & 66.3 & 69.4 & 52.4 \\
    \quad + T-GRPO                & 85.5 & 74.3 & \textbf{43.5} & \textbf{20.0} & 83.3 & 67.6 & 70.9 & 54.5 \\
    \quad + ArrowRL               & 84.2 & 71.2 & 42.0 & 19.4 & 82.1 & 64.6 & 69.6 & 52.2 \\
    \rowcolor{ourshade}
    \quad + \ours{CRPO (Ours)}    & \textbf{89.8} & \textbf{81.5} & 41.7 & 19.8 & \textbf{85.0} & \textbf{70.5} & \textbf{72.5}   & \textbf{58.1}   \\
    \medrule
    Random chance                 & 33.3 & 11.1 & 25.0 & 6.3  & 50.0 & 25.0 & 35.1 & 13.3 \\
    \thickrule
  \end{tabular}}
\end{table}

Three observations stand out. \textbf{(i)} CRPO yields its largest gains on Reversible Dynamics and Event Sequence, the two sub-tasks that most directly require temporal reasoning. In particular, Event Sequence is constructed through segment reordering, which is never used as a CRPO training transformation; CRPO reaches 62.7 P-Acc at 4B and 70.5 at 8B, compared with 57.0 and 67.6 for T-GRPO. \textbf{(ii)} Moving Direction remains challenging for all open-source models. T-GRPO is slightly higher than CRPO on this sub-task (21.8 vs.\ 21.0 P-Acc at 4B and 20.0 vs.\ 19.8 at 8B), showing that CRPO's gains are not uniform across every DyBench category. \textbf{(iii)} The much larger separation on P-Acc than on per-question Acc illustrates why paired evaluation is useful for diagnosing sensitivity to the underlying video dynamics.


\section{Training Data and Implementation Details}
\label{sec:app_train}

\subsection{Training data and reward functions}
\label{sec:app_data_reward}

\paragraph{Data sources.}
The training corpus combines text, image, and video QA, summarized in Table~\ref{tab:app_data}. The dual-branch counterfactual framework of CRPO is applied only to video samples, where spatial and temporal perturbations are meaningful. Text and image samples are trained with standard single-branch GRPO. Following standard practice, samples for which the base model is either always correct or always wrong across $8$ base-model rollouts are filtered out, since they carry no signal under group-relative advantages. The video subset contains $21{,}165$ samples after filtering, whose distribution across Task Router categories is reported in Appendix~\ref{sec:app_router}.

\begin{table}[htbp]
  \centering
  \caption{Training data composition.}
  \label{tab:app_data}
  \small
  \setlength{\tabcolsep}{8pt}
  \renewcommand{\arraystretch}{1.18}
  \arrayrulecolor{rulecolor}
  \begin{tabular}{l p{0.62\linewidth} r}
    \thickrule
    \textbf{Modality} & \textbf{Sources} & \textbf{Approx.\ size} \\
    \medrule
    Text  & DAPO-Math~\citep{dapo} & 6.4K \\
    Image & VIRL~\citep{wang2025vl}, ThinkLite-VL-Hard~\citep{wang2025sota} & 27.5K \\
    Video & Video-R1~\citep{videor1}, TVBench~\citep{cores2024lost}, STI-Bench~\citep{li2025sti}, MMR-VBench~\citep{zhu2025mmr} & 21.2K \\
    \thickrule
  \end{tabular}
\end{table}

\paragraph{Reward functions.}
We use two standard reward terms shared by all RL methods, namely a QA-style \emph{accuracy} reward and a \emph{format} reward. The accuracy reward is binary, with math problems verified by \texttt{math-verify} and other QA checked by case-folded, whitespace-stripped string match. The format reward is binary and enforces a strict regex of one $\langle\texttt{think}\rangle$ block followed by a single \texttt{\textbackslash boxed\{\}}. CRPO additionally introduces $R_{\text{CRR}}$ and $R_{\text{behave}}$ as defined in Section~\ref{sec:crr}.

\paragraph{Training prompt.}
All RL methods (CRPO and the baselines) use the following RL-with-Thinking system prompt during training.
\begin{promptbox}[Training System Prompt]
\small
You are a helpful assistant. \textbf{FIRST}, think through the reasoning process as an internal monologue, and \textbf{THEN} provide the final answer. The reasoning process MUST be enclosed within \texttt{<think>~</think>} tags, and the final answer MUST be wrapped in \texttt{\textbackslash boxed\{\}}.
\end{promptbox}
For multiple-choice questions, the option list is appended to the user prompt together with the appended ``None of the above'' (Appendix~\ref{app:null_option}).

\subsection{Hyperparameters}
\label{sec:app_train_script}

Table~\ref{tab:app_hparams} lists the key hyperparameters used by all RL methods (GRPO, T-GRPO, ArrowRL, and CRPO). Hyperparameters that differ between methods are limited to the CRPO-specific block at the bottom of the table.

\begin{table}[htbp]
  \centering
  \caption{Key training hyperparameters. The first block is shared by all RL baselines; the second block is CRPO-specific.}
  \label{tab:app_hparams}
  \small
  \setlength{\tabcolsep}{8pt}
  \renewcommand{\arraystretch}{1.18}
  \arrayrulecolor{rulecolor}
  \begin{tabular}{l l}
    \thickrule
    \textbf{Hyperparameter} & \textbf{Value} \\
    \medrule
    \groupheader{2}{Shared (all RL methods)} \\
    Backbone                       & Qwen3-VL-4B / 8B-Instruct~\citep{qwen3vl} \\
    Tunable modules                & LLM + multimodal projector (vision encoder frozen) \\
    Optimizer                      & AdamW (fused), $\beta_1{=}0.9$, $\beta_2{=}0.999$, weight decay $0.01$ \\
    Learning rate / schedule       & $1\!\times\!10^{-6}$, constant with warmup \\
    KL coefficient $\beta$         & $0.01$ \\
    Max gradient norm              & $1.0$ \\
    Batch size (per device)        & $4$  \\
    Generation steps per update    & $8$ \\
    Rollouts per group $G$         & $8$ \\
    Max prompt / completion length & $8192$ / $2048$ \\
    Sampled frames per video       & $32$ (uniform) \\
    Pixel budget per video         & $\le 2048{\times}32{\times}32$ \\
    Pixel budget per frame         & $[16,\,768]{\times}32{\times}32$ \\
    Epochs                         & $1$ \\
    Precision                      & BF16, ZeRO-3, gradient checkpointing \\
    Rollout backend                & vLLM (colocate, TP\,=\,8, GPU mem util.\ $0.45$) \\
    Hardware                       & 32 NVIDIA H20 GPUs (4 nodes $\times$ 8) \\
    \medrule
    \rowcolor{ourshade}
    \multicolumn{2}{l}{\textit{\textcolor{accentclr}{CRPO-specific}}} \\
    \rowcolor{ourshade}
    Cross-branch CRR weight $\lambda_d{=}\lambda_s$ & $0.3$ \\
    \rowcolor{ourshade}
    Counterfactual branch weight $w_{\text{aug}}$    & $0.5$ \\
    \rowcolor{ourshade}
    Counterfactual transformations                   & horizontal flip; temporal reversal \\
    \thickrule
  \end{tabular}
\end{table}

\paragraph{Evaluation toolkit.}
All evaluations are run with the public VLMEvalKit~\citep{vlmevalkit} toolkit\footnote{\url{https://github.com/open-compass/VLMEvalKit}} with its default configuration of 32 uniformly sampled frames per video for video benchmarks. We do not change any benchmark-side prompts or option ordering, and we use greedy decoding.

\subsection{Baseline RL algorithms}
\label{sec:app_baselines}

For a fair comparison, all RL post-training baselines (GRPO, T-GRPO, ArrowRL) are reproduced under the same backbone, training data, and shared hyperparameters as CRPO. They differ only in the RL algorithm. We reimplement each baseline strictly following its original paper.

\paragraph{GRPO~\citep{grpo}.}
The vanilla baseline. For each prompt, the policy generates $G{=}8$ rollouts, and each rollout receives the standard correctness reward $R_{\text{correct}} = \mathbb{I}[o = o^*]$ plus the format reward $R_{\text{format}}$. Advantages are computed by subtracting the group mean reward and normalizing by the group standard deviation, and the policy is updated with the clipped GRPO surrogate objective.

\paragraph{T-GRPO~\citep{videor1}.}
Following Video-R1, for each prompt we additionally generate a second group of $G{=}8$ rollouts on a frame-shuffled version of the input video. Let $p$ and $\tilde{p}$ denote the fraction of correct rollouts in the ordered and shuffled groups, respectively. We add a temporal reward $r_t = \alpha \cdot \mathbb{I}[p \ge \tilde{p}]$ with $\alpha = 0.3$ to each correct rollout in the ordered group. Following the original paper, the shuffled group is used only to compute $\tilde{p}$ and does \emph{not} contribute to the policy gradient, so the optimization is effectively single-branch. The temporal reward is applied only to video samples.

\paragraph{ArrowRL~\citep{xue2025seeing}.}
Following the original paper, for each prompt we generate one rollout group on the original video and a single reference response $\tilde{o}$ on the temporally reversed video. The reward for each original-video rollout $o_i$ combines a fidelity term and a reverse term, $r_i = r_i^{\text{fid}} + \alpha_i \cdot r_i^{\text{rev}}$, where $r_i^{\text{fid}} = \mathbb{I}[o_i = o^*]$ for multiple-choice questions, $r_i^{\text{rev}} = 1 - \mathbb{I}[o_i=\tilde{o}]$ penalizes rollouts that mirror the reverse-conditioned response, and $\alpha_i$ is a dynamic weight set to $\alpha = 0.25$ for AoT-sensitive samples ($\tilde{o}\neq o^*$) and $0$ otherwise ($\tilde{o}= o^*$). The reversed video is used only to compute $\tilde{o}$ for reward shaping and does not contribute policy gradients.


\section{Task Router}
\label{sec:app_router}
\label{app:router}

\paragraph{Classification model.}
The Task Router is run \emph{offline} once on the entire training set with DeepSeek-R1, a text-only reasoning model. The classification is text-only and therefore does not require access to the video itself.

\paragraph{Prompt template.}
The full prompt template is reproduced below. It is the verbose implementation of the two imagination questions shown in Figure~\ref{fig:crpo} (right panel), with detailed criteria for what counts as a YES versus a NO under each test. Since the Task Router classifies only video samples, the prompt is designed around video-specific transformations (horizontal flip and temporal reversal).

\begin{promptbox}[Task Router Prompt (used with DeepSeek-R1)]
\small
\textbf{Task.} Classify this VQA item into [\texttt{Temporal}, \texttt{Spatial}, \texttt{Spatiotemporal}, \texttt{Static}] based on two transformation tests.

\medskip
\textbf{1.~Horizontal Flip Test (Spatial Sensitivity).} If the video is mirrored horizontally (Left $\leftrightarrow$ Right):
\begin{itemize}\setlength\itemsep{0pt}
  \item \textcolor{green!50!black}{\textbf{YES (Change)}}: queries about left/right, clockwise/counter-clockwise, or horizontal orientation.
  \item \textcolor{red!70!black}{\textbf{NO (No Change)}}: queries about verticality (top/bottom, high/low, tall/short), color, identity, size, or count.
\end{itemize}

\textbf{2.~Time-Reversal Test (Temporal Sensitivity).} If the video is played in REVERSE (frames played backwards, time flows backward):
\begin{itemize}\setlength\itemsep{0pt}
  \item \textcolor{green!50!black}{\textbf{YES (Change)}}: directional actions where forward $\neq$ backward (opening$\to$closing, picking up$\to$putting down, entering$\to$exiting); questions about FIRST/LAST or BEFORE/AFTER; questions about cause/result of a sequence.
  \item \textcolor{red!70!black}{\textbf{NO (No Change)}}: counting, identity/color/material/shape/size, persistent states or attributes, motion speed that are symmetric under reversal.
\end{itemize}

\textbf{Decision Matrix.}
\begin{itemize}\setlength\itemsep{0pt}
  \item \texttt{Spatial}: Flip \textbf{YES} \textsc{and} Time-Reversal \textbf{NO}.
  \item \texttt{Temporal}: Flip \textbf{NO} \textsc{and} Time-Reversal \textbf{YES}.
  \item \texttt{Spatiotemporal}: \textbf{both YES}.
  \item \texttt{Static}: \textbf{both NO}.
\end{itemize}

\textbf{Data.} Q: \texttt{\{problem\}}\quad Options: \texttt{\{options\}}\quad A: \texttt{\{solution\}}

\medskip
\textbf{Instructions.} First, briefly reason through the two tests. Then, on the very last line, output ONLY one word from $\{$\texttt{Temporal}, \texttt{Spatial}, \texttt{Spatiotemporal}, \texttt{Static}$\}$.
\end{promptbox}

\paragraph{Statistics on the training set.}
Table~\ref{tab:app_router_stats} reports the per-source distribution. The video subset is dominated by \texttt{Temporal} questions (73.6\%), reflecting that most existing video QA data is action- or event-centric. Pure \texttt{Spatial} questions are rare (2.3\%), and \texttt{Spatiotemporal} questions, which require both axes, account for 4.1\%. The \texttt{Static} category (20.0\%) covers static questions about presence, count, color, etc.

\begin{table}[htbp]
  \centering
  \caption{Task Router classification statistics on the video training set (21{,}165 samples).}
  \label{tab:app_router_stats}
  \small
  \setlength{\tabcolsep}{6pt}
  \renewcommand{\arraystretch}{1.18}
  \arrayrulecolor{rulecolor}
  \begin{tabular}{l r r r r r}
    \thickrule
    \textbf{Dataset} & \textbf{Temporal} & \textbf{Spatial} & \textbf{Spatiotemp.} & \textbf{Static} & \textbf{Total} \\
    \medrule
    Video-R1   & 14{,}359 (80.8\%) & 173 ( 1.0\%) & 310 ( 1.7\%) & 2{,}936 (16.5\%) & 17{,}778 \\
    TVBench    &    979 (70.9\%) &   2 ( 0.1\%) & 238 (17.2\%) &   161 (11.7\%) &  1{,}380 \\
    STI-Bench  &     81 ( 5.3\%) & 313 (20.5\%) & 319 (20.8\%) &   817 (53.4\%) &  1{,}530 \\
    MMR-VBench &    163 (34.2\%) &   3 ( 0.6\%) &   1 ( 0.2\%) &   310 (65.0\%) &     477 \\
    \medrule
    \rowcolor{ourshade}
    \textbf{Total} & \textbf{15{,}582 (73.6\%)} & \textbf{491 (2.3\%)} & \textbf{868 (4.1\%)} & \textbf{4{,}224 (20.0\%)} & \textbf{21{,}165} \\
    \thickrule
  \end{tabular}
\end{table}

\paragraph{Validation by manual relabeling.}
To verify the router's quality, we randomly sample 200 video QA examples and manually relabel them by inspecting the video, question, and options. The router agrees with the human labels on 188 of 200 cases (\textbf{94\%}). Eight of the 12 disagreements are borderline \texttt{Temporal}/\texttt{Spatiotemporal} cases, which still select temporal reversal as a valid transformation; the remaining four involve \texttt{Static} or \texttt{Spatial} mismatches.

\paragraph{Robustness to routing errors.}
\label{app:router_robustness}
The manual audit above measures the Task Router's prediction quality, but does
not directly reveal how sensitive CRPO is to routing errors.
We therefore conduct a controlled noise experiment on Qwen3-VL-4B, randomly
corrupting the router assignments of $10\%$ or $20\%$ of the video training
samples while keeping all other training settings unchanged.
Table~\ref{tab:router_noise} reports the results.
A $10\%$ corruption rate causes only small changes, reducing DyBench P-Acc from
$54.8$ to $54.2$ and TimeBlind I-Acc from $31.7$ to $31.3$.
Performance degrades more clearly at $20\%$ corruption, but remains above the
Qwen3-VL-4B base model on both paired benchmarks.
Together with the $94\%$ agreement in the manual audit, these results indicate
that CRPO does not require a perfectly accurate router.

\begin{table}[htbp]
    \centering
    \caption{\textbf{Robustness to Task Router errors on Qwen3-VL-4B.}
    We randomly corrupt the router assignments for a fraction of video training
    samples and retrain CRPO with all other settings unchanged.}
    \label{tab:router_noise}
    \small
    \setlength{\tabcolsep}{5.5pt}
    \renewcommand{\arraystretch}{1.12}
    \arrayrulecolor{rulecolor}
    \begin{tabular}{lcccc}
        \thickrule
        \textbf{Router noise}
        & \multicolumn{2}{c}{\textbf{DyBench}}
        & \multicolumn{2}{c}{\textbf{TimeBlind}} \\
        \cmidrule(lr){2-3}
        \cmidrule(lr){4-5}
        & Acc & P-Acc & Acc & I-Acc \\
        \medrule
        0\%  & 70.3 & 54.8 & 69.8 & 31.7 \\
        10\% & 70.0 & 54.2 & 69.6 & 31.3 \\
        20\% & 68.6 & 51.4 & 67.9 & 29.4 \\
        \thickrule
    \end{tabular}
\end{table}

\begin{figure}[htbp]
  \centering
  \includegraphics[width=\textwidth]{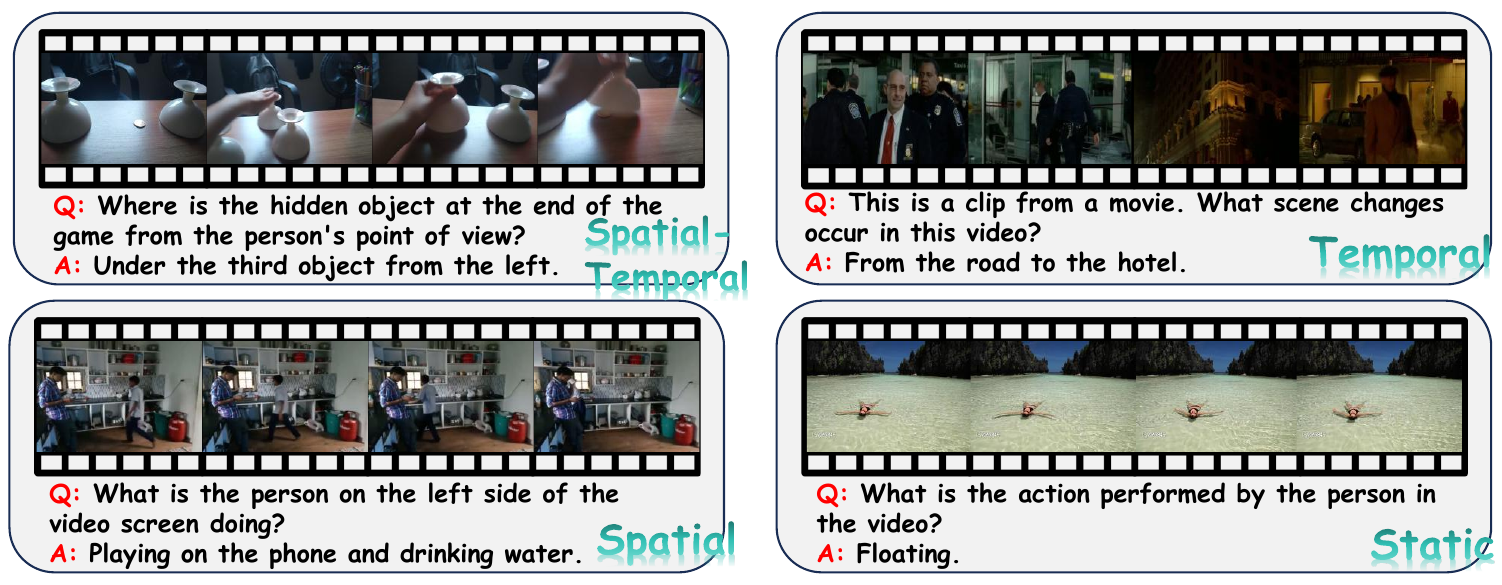}
  \caption{\textbf{Task Router output examples} after applying the prompt above to randomly sampled videos. The router outputs a category label together with a short text-only reasoning chain.}
  \label{fig:task_example}
\end{figure}

\paragraph{Example router outputs.}
Figure~\ref{fig:task_example} shows representative router outputs across the four categories, including the question and ground-truth answer.

\begin{figure}[ht]
  \centering
  \includegraphics[width=0.6\textwidth]{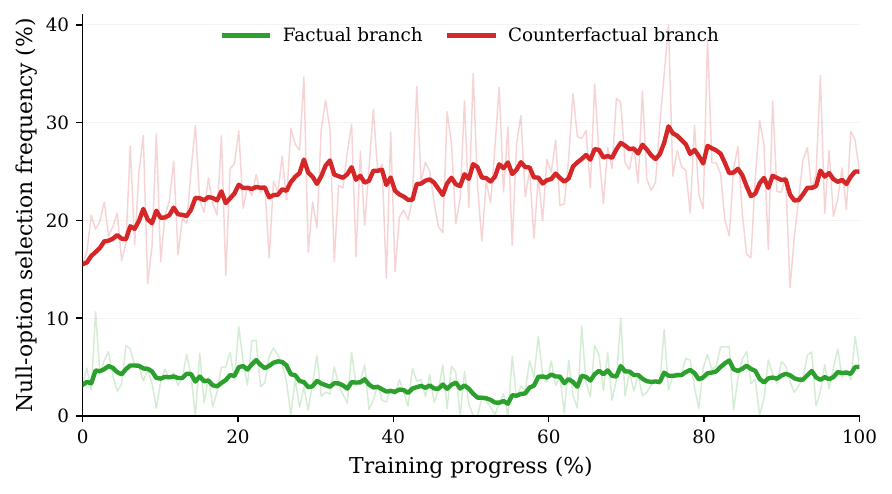}
  \caption{\textbf{Null-Option selection frequency during Qwen3-VL-4B
  training.}
  Thin curves show per-batch frequencies and thick curves show smoothed trends.
  Null-Option usage remains low on factual rollouts, while counterfactual
  rollouts fluctuate at a substantially higher level and stabilize around
  $25\%$ rather than increasing continuously.}
  \label{fig:null_frequency}
\end{figure}

\section{Additional CRPO Implementation Details}
\label{sec:app_method}

\subsection{Null Option}
\label{app:null_option}

For multiple-choice training examples, a flip or temporal reversal can move the
correct counterfactual answer outside the original answer choices.
We therefore append a \emph{None of the above} option
$o^{\mathcal N}$ during RL training.
On the factual branch, the original ground-truth answer remains present, so the
Null Option is always incorrect.
On a dynamic counterfactual branch, it allows the policy to express that none
of the original choices remains valid after the intervention; on Static
questions it remains incorrect because the answer should be preserved.
The Null Option is used only during training and is not added to benchmark
evaluations.
Its contribution is quantified by the ``w/o Null Option'' row in
Table~\ref{tab:ablation}.

\paragraph{Selection dynamics.}
\label{app:null_frequency}
A potential concern is that the policy may exploit the Null Option by selecting
it indiscriminately on counterfactual inputs.
Figure~\ref{fig:null_frequency} tracks its per-batch selection frequency during
Qwen3-VL-4B training.
On the factual branch, the frequency fluctuates across batches but remains low
on average, with the smoothed trend staying around a few percent throughout
training.
In contrast, the counterfactual branch exhibits substantially higher usage:
per-batch frequencies vary roughly between $12\%$ and $40\%$, while the
smoothed trend stabilizes around $25\%$ after the early stage of training.
Importantly, the frequency does not continue to increase with training, but
instead fluctuates around a stable level.
This branch-specific pattern is consistent with the intended role of the Null
Option: counterfactual transformations can invalidate the original answer
choices, whereas factual inputs retain an explicit valid target.
Together with factual correctness supervision and the invariance constraint on
Static examples, this behavior argues against a degenerate always-Null strategy.
The ablation in Table~\ref{tab:ablation} further shows that removing the Null
Option substantially reduces paired performance.

\subsection{Exact advantage normalization}
\label{app:optimization}
Let $R_i^{\mathrm F}$ and $R_j^{\mathrm{CF}}$ be the factual and counterfactual rewards from Eqs.~\ref{eq:factual_reward}--\ref{eq:counterfactual_reward}. CRPO first centers the two rollout groups separately,
\begin{align}
\delta_i^{\mathrm F} &= R_i^{\mathrm F} - \frac{1}{G}\sum_{k=1}^{G}R_k^{\mathrm F}, \\
\delta_j^{\mathrm{CF}} &= R_j^{\mathrm{CF}} - \frac{1}{G}\sum_{k=1}^{G}R_k^{\mathrm{CF}}.
\end{align}
It then computes a shared scale over the $2G$ centered rewards,
\begin{equation}
\sigma_{\mathrm{shared}}
=
\sqrt{
\frac{1}{2G}
\left(
\sum_{i=1}^{G}(\delta_i^{\mathrm F})^2
+
\sum_{j=1}^{G}(\delta_j^{\mathrm{CF}})^2
\right)
+\epsilon
},
\end{equation}
and uses
\begin{equation}
\widehat A_i^{\mathrm F}=\frac{\delta_i^{\mathrm F}}{\sigma_{\mathrm{shared}}},
\qquad
\widehat A_j^{\mathrm{CF}}=\frac{\delta_j^{\mathrm{CF}}}{\sigma_{\mathrm{shared}}}.
\end{equation}
Thus, each branch preserves its own within-group ranking while both branches are placed on the same optimization scale. We apply the standard GRPO token-level KL regularizer to the reference policy on both branches, using the same KL coefficient $\beta=0.01$ as in the controlled baselines (Table~\ref{tab:app_hparams}).

\subsection{Auxiliary-reward cancellation under group-wise normalization}
\label{app:cancellation}
The cancellation discussed in Section~\ref{sec:cancellation} arises in the common case where a perturbation-derived bonus acts as the same multiplicative factor on all correct rollouts in a group. Ignoring the format term for clarity, let the binary correctness reward be $c_i\in\{0,1\}$ and let a group-level signal produce $R_i=(1+C)c_i$. If $\bar c$ and $\sigma_c$ are the mean and standard deviation of $c_i$, then the normalized advantage is
\begin{equation}
\widehat A_i
=
\frac{(1+C)(c_i-\bar c)}{(1+C)\sigma_c+\epsilon}
\approx
\frac{c_i-\bar c}{\sigma_c},
\end{equation}
where the approximation becomes exact as the numerical stabilizer becomes negligible. The auxiliary factor therefore contributes little or no new relative advantage in this case. This analysis does not imply that every perturbation-based reward is always cancelled; it identifies a structural failure mode when the auxiliary signal is shared across the rollout group.

CRPO receives a qualitatively different signal because the counterfactual branch generates independent rollouts with per-rollout behavioral rewards $b_j$. Some counterfactual rollouts satisfy the expected response and others do not, producing within-group variation that survives normalization and directly contributes policy gradients. In addition, CRPO's shared cross-branch standard deviation prevents each branch from independently rescaling away its cross-branch reward factor. These two properties explain why counterfactual information enters CRPO through the optimized rollouts themselves rather than only through a group-level reward modifier.

\section{Additional Training Analysis}
\label{sec:app_training_analysis}

\subsection{The four CRR reward components}
\begin{figure}[htbp]
  \centering
  \includegraphics[width=0.95\textwidth]{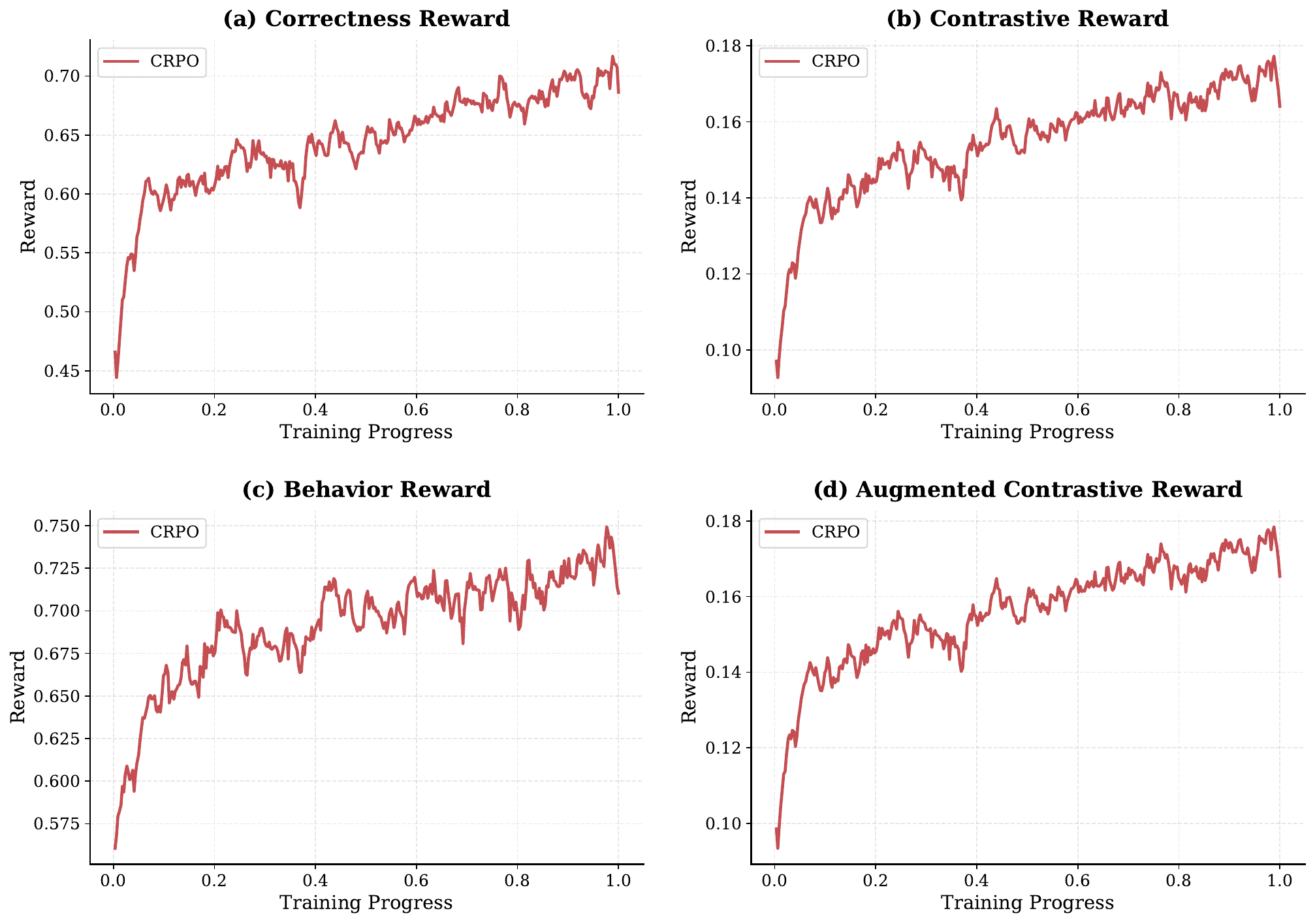}
  \caption{\textbf{The four reward components introduced by CRPO.}}
  \label{fig:crpo_rewards}
\end{figure}

Figure~\ref{fig:crpo_rewards} visualizes the four reward terms introduced by CRPO. The original branch combines a correctness reward $R_{\text{correct}}$ and a cross-branch CRR reward $R_{\text{CRR}}^{\text{orig}}$ that scores the counterfactual rollouts. The counterfactual branch combines a behavioral reward $R_{\text{behave}}$ and a cross-branch CRR reward $R_{\text{CRR}}^{\text{aug}}$ that scores the original rollouts. Both branches additionally use the inherited format reward.

A natural observation from Figure~\ref{fig:crpo_rewards} is that the two cross-branch curves $R_{\text{CRR}}^{\text{orig}}$ and $R_{\text{CRR}}^{\text{aug}}$ track each other almost perfectly throughout training. This is not a plotting artifact but a direct consequence of the symmetric mutual-reward design. Let $p$ denote the fraction of correct rollouts in the original branch and $q$ the fraction of behavior-matching rollouts in the counterfactual branch. By construction, $R_{\text{CRR}}^{\text{orig}}$ activates only on the $pG$ correct original rollouts and each of them measures how often the counterfactual rollouts behave as expected, giving a batch-level expectation of $\lambda \cdot p \cdot q$. The counterfactual side is analogous: $R_{\text{CRR}}^{\text{aug}}$ activates only on the $qG$ behavior-matching counterfactual rollouts and each measures the original branch's correctness rate $p$, again yielding $\lambda \cdot p \cdot q$. The two CRR streams therefore receive signals of equal magnitude and neither branch's reward dominates the other.

\subsection{Training dynamics on Qwen3-VL-8B}
\label{sec:app_reward_8b}

\begin{figure}[htbp]
  \centering
  \includegraphics[width=\textwidth]{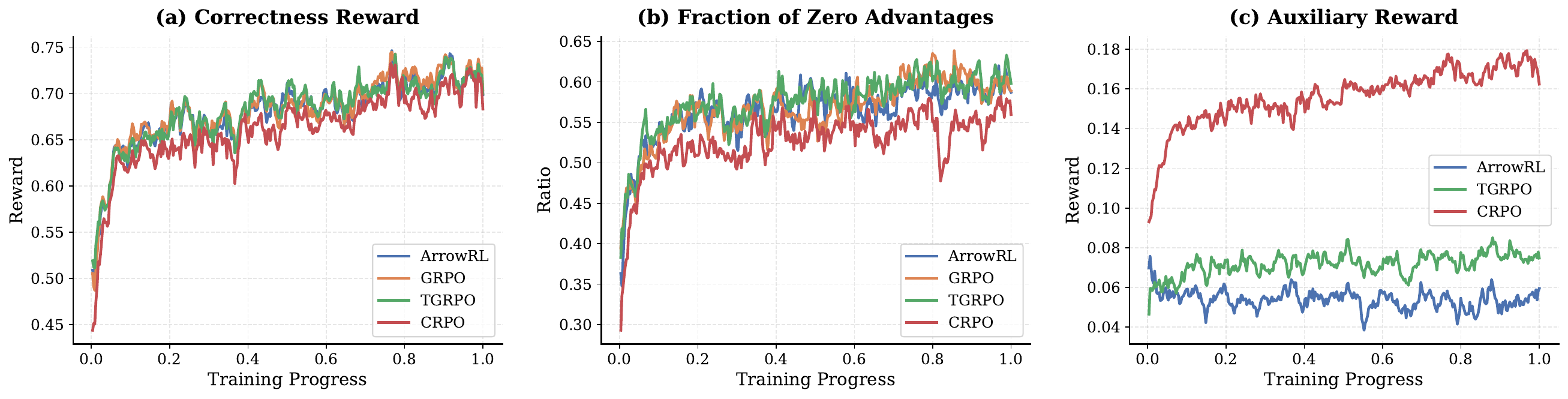}
  \caption{\textbf{Training dynamics of CRPO vs.\ RL baselines on Qwen3-VL-8B.} Same layout as Figure~\ref{fig:reward} (4B). The trends are consistent across model scales.}
  \label{fig:reward_8b}
\end{figure}

Figure~\ref{fig:reward_8b} plots the same three training-curve diagnostics as Figure~\ref{fig:reward} in the main paper for Qwen3-VL-8B. The trends mirror the 4B results: CRPO's factual correctness rises more gradually before converging to a comparable level, the fraction of zero-advantage groups remains lower, and the relational reward grows during training while the perturbation-derived rewards of T-GRPO and ArrowRL remain relatively flat. The similarity across the two backbone sizes indicates that the observed optimization behavior is not specific to the 4B model.

\subsection{Sensitivity to reward weights}
\label{sec:app_sensitivity}

To probe how robust CRPO is to its two reward coefficients, we run two one-dimensional sweeps on Qwen3-VL-4B and report two complementary metrics, the strict paired metric \textbf{DyBench P-Acc} (which most directly reflects spatiotemporal sensitivity) and the standard accuracy \textbf{VideoMME} (which reflects general video understanding).

\begin{itemize}\setlength\itemsep{0pt}
  \item \textbf{Sweep~A.} Vary the cross-branch CRR weight $\lambda{=}\lambda_d{=}\lambda_s$ over $\{0.0,\,0.1,\,0.3,\,0.5\}$ with $w_{\text{aug}}{=}0.5$ fixed. The point $\lambda{=}0$ corresponds to a dual-branch GRPO without any CRR signal, isolating the contribution of the relational reward.
  \item \textbf{Sweep~B.} Vary the counterfactual-branch weight $w_{\text{aug}}$ over $\{0.0,\,0.3,\,0.5,\,0.7\}$ with $\lambda{=}0.3$ fixed. The point $w_{\text{aug}}{=}0$ disables direct policy updates from the counterfactual branch while retaining the counterfactual rollouts used to compute the factual-branch relational reward.
\end{itemize}

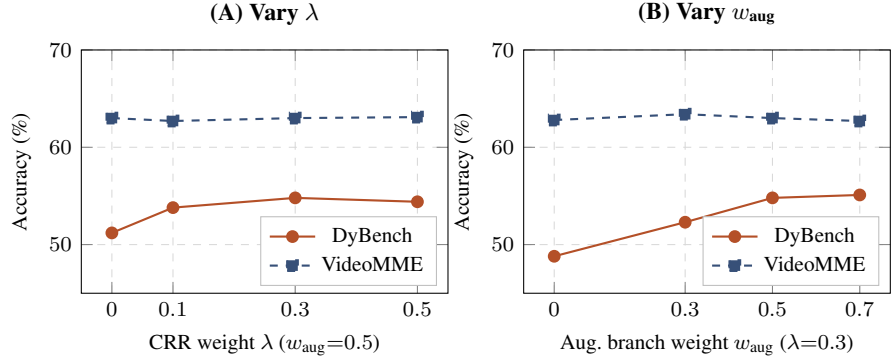
\begin{figure}[htbp]
  \centering
  \begin{tikzpicture}
    \begin{axis}[
      name=axA,
      width=0.46\textwidth, height=4.8cm,
      xlabel={CRR weight $\lambda$ ($w_{\text{aug}}{=}0.5$)},
      ylabel={Accuracy (\%)},
      xtick={0,0.1,0.3,0.5},
      ymin=45, ymax=70,
      grid=major, grid style={dashed,gray!30},
      legend pos=south east,
      legend style={font=\footnotesize, draw=gray!50},
      tick label style={font=\footnotesize},
      label style={font=\footnotesize},
      title style={font=\small\bfseries},
      title={(A) Vary $\lambda$},
    ]
      \addplot[mark=*, thick, color=accentclr] coordinates {
        (0.0, 51.2)
        (0.1, 53.8)
        (0.3, 54.8)   
        (0.5, 54.4)
      };
      \addlegendentry{DyBench}
      \addplot[mark=square*, thick, color=groupclr, dashed] coordinates {
        (0.0, 63.0)
        (0.1, 62.7)
        (0.3, 63.0)   
        (0.5, 63.1)
      };
      \addlegendentry{VideoMME}
    \end{axis}

    \begin{axis}[
      at={(axA.south east)}, xshift=1.0cm, anchor=south west,
      width=0.46\textwidth, height=4.8cm,
      xlabel={Aug.\ branch weight $w_{\text{aug}}$ ($\lambda{=}0.3$)},
      ylabel={Accuracy (\%)},
      xtick={0,0.3,0.5,0.7},
      ymin=45, ymax=70,
      grid=major, grid style={dashed,gray!30},
      legend pos=south east,
      legend style={font=\footnotesize, draw=gray!50},
      tick label style={font=\footnotesize},
      label style={font=\footnotesize},
      title style={font=\small\bfseries},
      title={(B) Vary $w_{\text{aug}}$},
    ]
      \addplot[mark=*, thick, color=accentclr] coordinates {
        (0.0, 48.8)
        (0.3, 52.3)
        (0.5, 54.8)   
        (0.7, 55.1)
      };
      \addlegendentry{DyBench}
      \addplot[mark=square*, thick, color=groupclr, dashed] coordinates {
        (0.0, 62.8)
        (0.3, 63.4)
        (0.5, 63.0)   
        (0.7, 62.7)
      };
      \addlegendentry{VideoMME}
    \end{axis}
  \end{tikzpicture}
  \caption{\textbf{Sensitivity of CRPO to its two reward coefficients} on Qwen3-VL-4B. Solid orange: DyBench P-Acc (spatiotemporal-sensitive); dashed navy: VideoMME (general). \emph{Left:} varying the cross-branch CRR weight $\lambda$ with $w_{\text{aug}}{=}0.5$ fixed. \emph{Right:} varying the counterfactual-branch weight $w_{\text{aug}}$ with $\lambda{=}0.3$ fixed.}
  \label{fig:sensitivity}
\end{figure}

Figure~\ref{fig:sensitivity} plots the results.

\textbf{Takeaways.} Setting $\lambda{=}0$ lowers DyBench P-Acc from 54.8 to 51.2, showing that the cross-branch relational term contributes beyond the local branch rewards. Setting $w_{\text{aug}}{=}0$ produces a larger drop to 48.8, consistent with the main ablation showing that direct policy updates from counterfactual rollouts are especially important. Across the non-zero settings, DyBench P-Acc changes modestly, while VideoMME remains within a narrow range, indicating that CRPO does not depend on a finely tuned coefficient choice.

\section{Additional Qualitative Examples}
\label{sec:app_examples}

\begin{figure}[htbp]
  \centering
  \includegraphics[width=\textwidth]{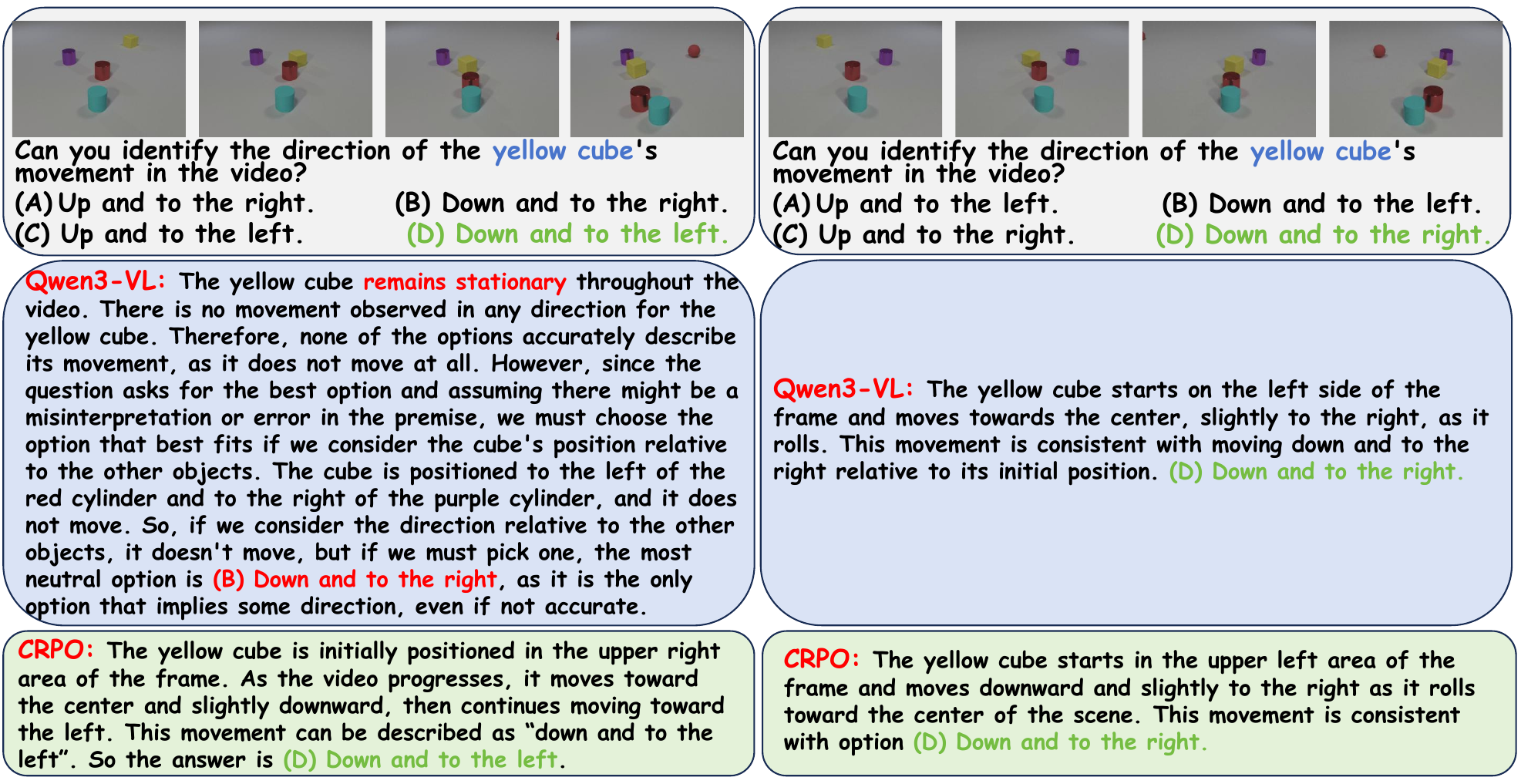}
  \caption{Qualitative example from DyBench (\emph{moving direction}): identifying which way the yellow cube moves on opposite paired videos.}
  \label{fig:example1}
\end{figure}

\begin{figure}[htbp]
  \centering
  \includegraphics[width=\textwidth]{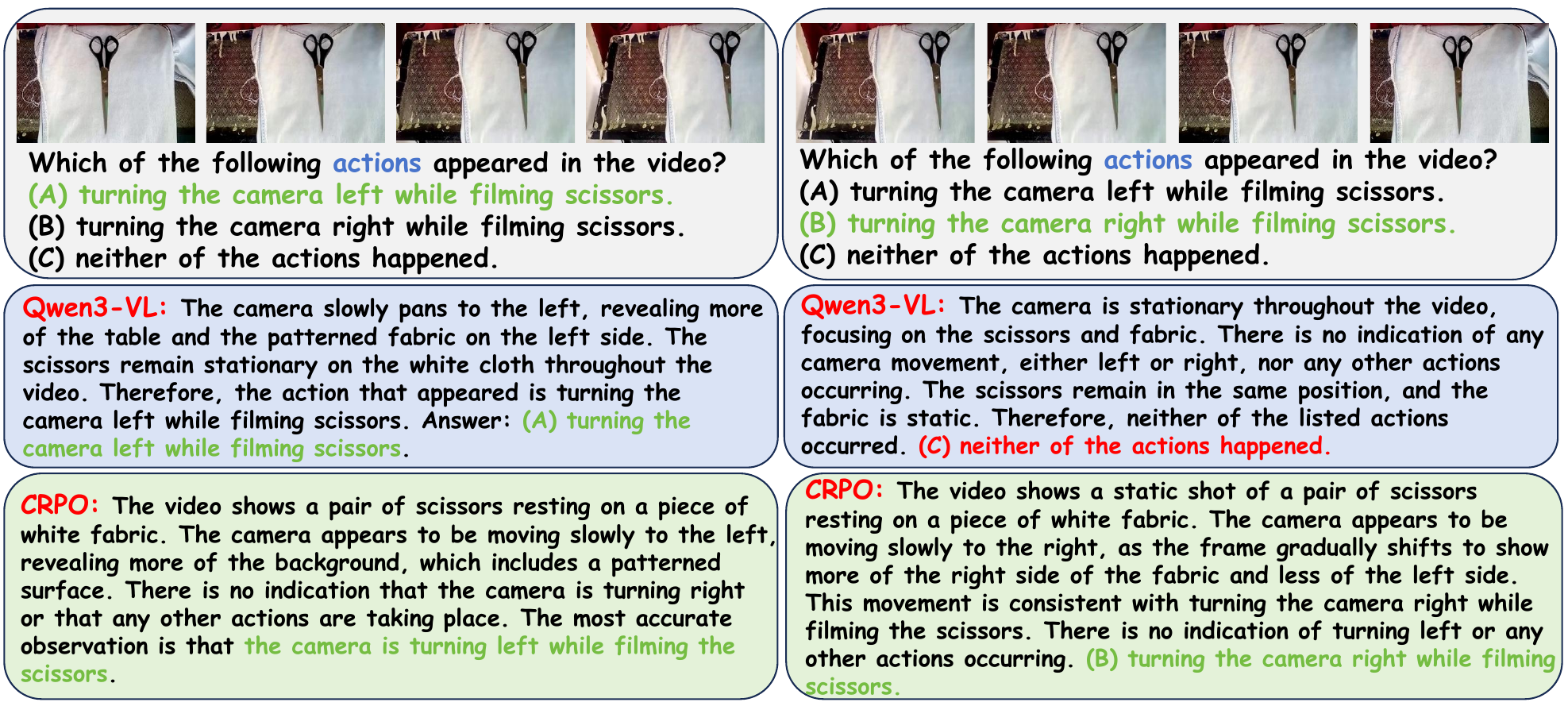}
  \caption{Qualitative example from DyBench (\emph{moving direction}): identifying whether the camera pans left or right while filming the scissors.}
  \label{fig:example2}
\end{figure}

\begin{figure}[htbp]
  \centering
  \includegraphics[width=\textwidth]{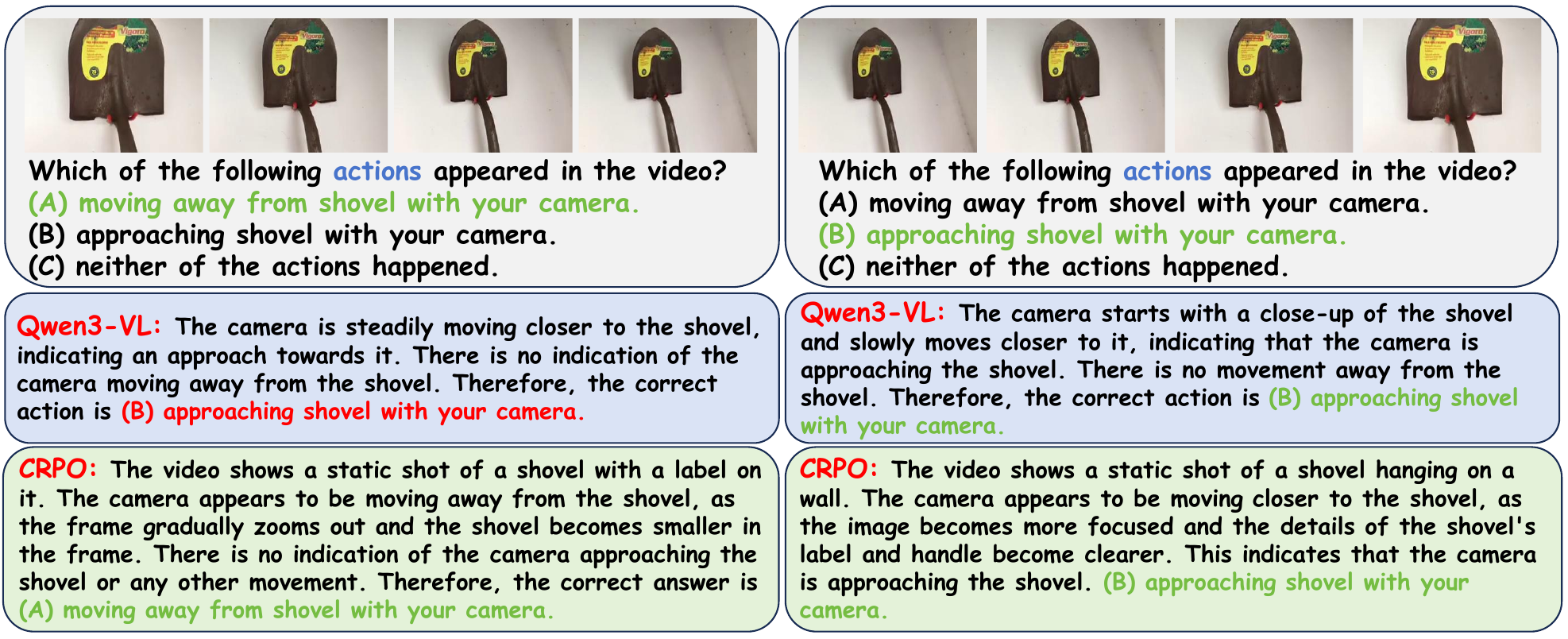}
  \caption{Qualitative example from DyBench (\emph{moving direction}): identifying whether the camera is approaching or moving away from the shovel.}
  \label{fig:example3}
\end{figure}

\begin{figure}[htbp]
  \centering
  \includegraphics[width=\textwidth]{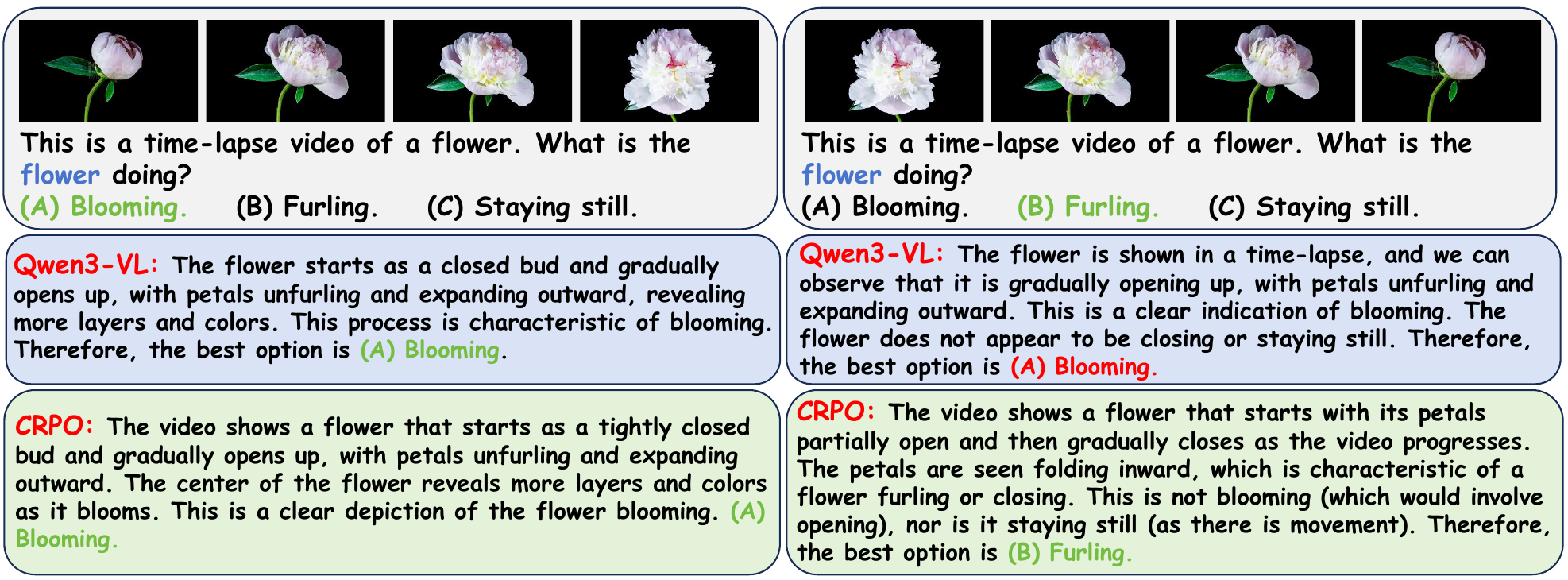}
  \caption{Qualitative example from DyBench (\emph{reversible dynamics}): identifying whether a time-lapse of a flower shows blooming or furling.}
  \label{fig:example4}
\end{figure}

\begin{figure}[htbp]
  \centering
  \includegraphics[width=\textwidth]{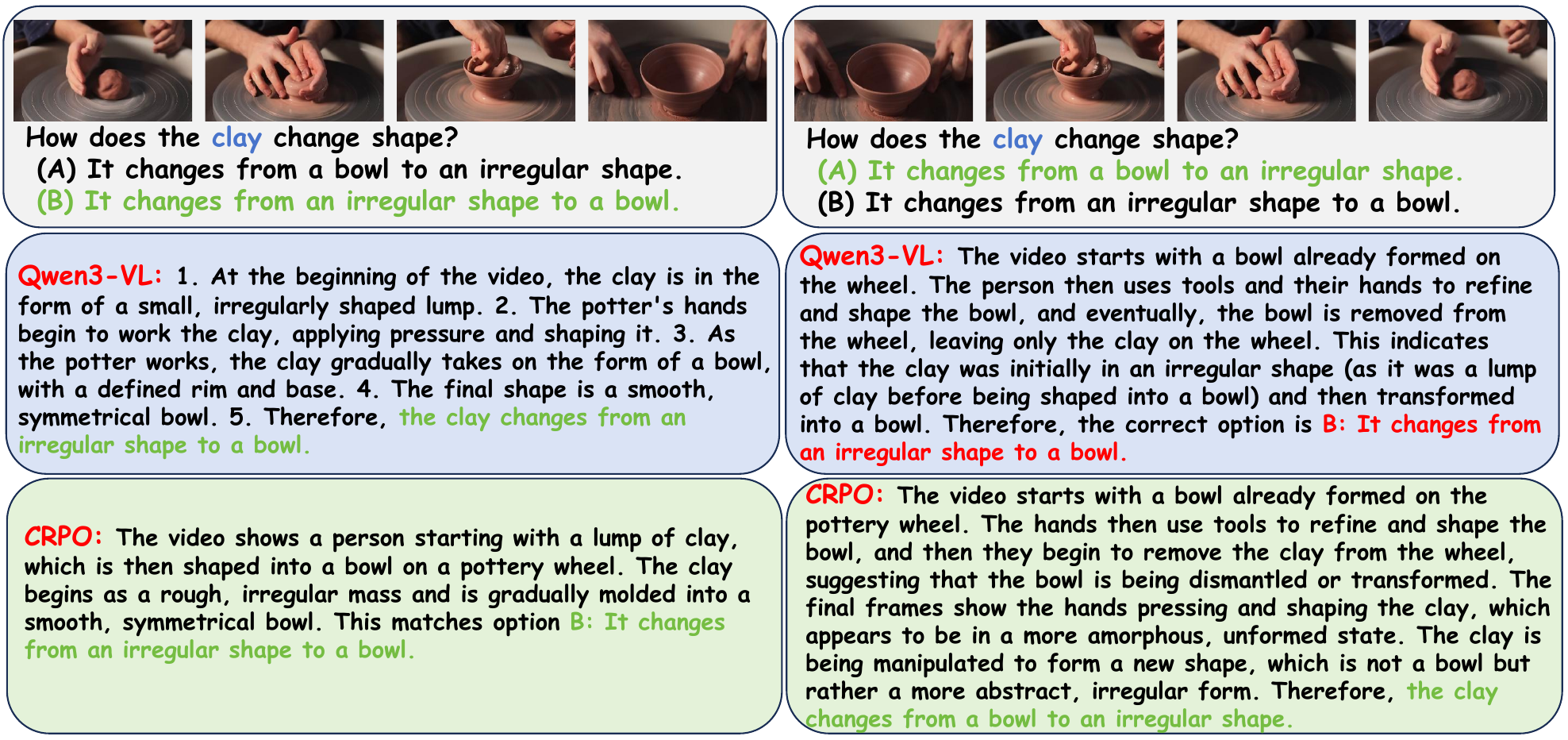}
  \caption{Qualitative example from TimeBlind: identifying whether the clay changes from a lump to a bowl, or from a bowl to a lump.}
  \label{fig:example5}
\end{figure}

\begin{figure}[htbp]
  \centering
  \includegraphics[width=\textwidth]{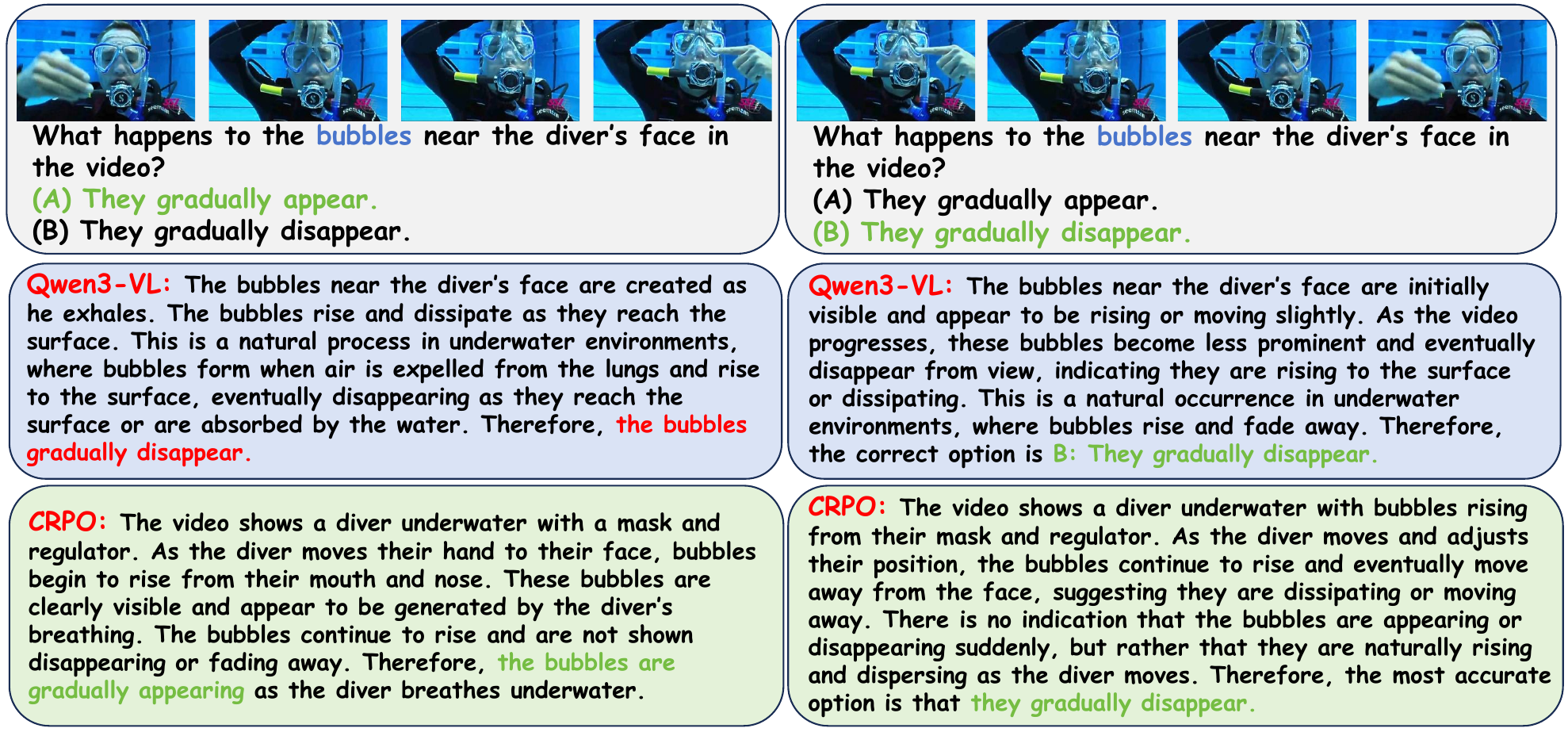}
  \caption{Qualitative example from TimeBlind: identifying whether the bubbles near the diver's face gradually appear or gradually disappear.}
  \label{fig:example6}
\end{figure}

Figures~\ref{fig:example1}--\ref{fig:example6} show additional qualitative comparisons between Qwen3-VL (baseline) and CRPO. The first four examples (Figures~\ref{fig:example1}--\ref{fig:example4}) are drawn from \textbf{DyBench}, while the last two (Figures~\ref{fig:example5}--\ref{fig:example6}) are drawn from \textbf{TimeBlind}~\citep{timeblind}. Across these examples, CRPO answers both sides of each pair correctly, while the baseline either repeats the same answer across the pair or predicts an incorrect direction or event order.

\section{Limitations and Broader Impact}
\label{sec:app_limit}

\paragraph{Limitations.}
CRPO currently uses horizontal flipping and temporal reversal as broadly applicable counterfactual transformations. Extending the framework to richer semantic interventions, such as object substitution or state editing, would require reliable video editing or curated counterfactual data. The Task Router also introduces a one-time offline preprocessing step.

\paragraph{Broader impact.}
CRPO is intended to make Video LLM predictions more responsive to the spatiotemporal evidence in the input, which may improve reliability in applications that depend on motion direction or event order. It does not eliminate model errors or provide guarantees for safety-critical use, and the broader dual-use considerations remain those of Video LLMs and multimodal foundation models more generally.

\end{document}